\documentclass[letterpaper]{article} 
\usepackage{aaai25}  
\usepackage{times}  
\usepackage{helvet}  
\usepackage{courier}  
\usepackage[hyphens]{url}  
\usepackage{graphicx} 
\usepackage{natbib}  
\usepackage{caption} 
\usepackage{algorithm}
\usepackage{algorithmicx}
\usepackage{algpseudocodex}

\usepackage{newfloat}
\usepackage{listings}
\DeclareCaptionStyle{ruled}{labelfont=normalfont,labelsep=colon,strut=off} 
\floatstyle{ruled}
\newfloat{listing}{tb}{lst}{}
\floatname{listing}{Listing}
\usepackage{amssymb}
\usepackage{amsmath}
\DeclareMathOperator*{\argmax}{arg\,max}

\usepackage{subcaption}
\usepackage{booktabs}
\usepackage{pifont}
\usepackage{amsfonts}
\usepackage{atveryend}
\usepackage{listings}
\usepackage{xcolor}
\definecolor{mydarkblue}{rgb}{0,0.08,0.45}
\usepackage[
    colorlinks=true,
    linkcolor=mydarkblue,
    citecolor=mydarkblue,
    filecolor=mydarkblue,
    urlcolor=mydarkblue,
]{hyperref}
\usepackage[capitalise]{cleveref}

\lstdefinelanguage{json}{
    basicstyle=\ttfamily\scriptsize, 
    numberstyle=\scriptsize,
    stepnumber=1,
    numbersep=8pt,
    showstringspaces=false,
    breaklines=true,
    frame=lines,
    backgroundcolor=\color{gray!10},
    tabsize=2, 
    literate=
     *{0}{{{\color{blue}0}}}{1}
      {1}{{{\color{blue}1}}}{1}
      {2}{{{\color{blue}2}}}{1}
      {3}{{{\color{blue}3}}}{1}
      {4}{{{\color{blue}4}}}{1}
      {5}{{{\color{blue}5}}}{1}
      {6}{{{\color{blue}6}}}{1}
      {7}{{{\color{blue}7}}}{1}
      {8}{{{\color{blue}8}}}{1}
      {9}{{{\color{blue}9}}}{1}
      {:}{{{\color{red}:}}}{1}
      {,}{{{\color{red},}}}{1}
      {\{}{{{\color{brown}\{}}}{1}
      {\}}{{{\color{brown}\}}}}{1}
      {[}{{{\color{brown}[}}}{1}
      {]}{{{\color{brown}]}}}{1},
}

\usepackage{xspace}

\makeatletter
\DeclareRobustCommand\onedot{\futurelet\@let@token\@onedot}
\def\@onedot{\ifx\@let@token.\else.\null\fi\xspace}

\def\eg{\emph{e.g}\onedot} 
\def\ie{\emph{i.e}\onedot} 
\def\cf{\emph{c.f}\onedot} 
 
\def\wrt{w.r.t\onedot} 

\makeatother

\title{Does VLM Classification Benefit from LLM Description Semantics?}
\author{
    Pingchuan Ma\equalcontrib,
    Lennart Rietdorf\equalcontrib,\\
    Dmytro Kotovenko,
    Vincent Tao Hu,
    Björn Ommer
}

\affiliations{
    CompVis @ LMU Munich \\
    Munich Center for Machine Learning \\
}

\begin{document}
\maketitle

\begin{abstract}
Accurately describing images with text is a foundation of explainable AI. Vision-Language Models (VLMs) like CLIP have recently addressed this by aligning images and texts in a shared embedding space, expressing semantic similarities between vision and language embeddings. VLM classification can be improved with descriptions generated by Large Language Models (LLMs). However, it is difficult to determine the contribution of actual description semantics, as the performance gain may also stem from a semantic-agnostic ensembling effect, where multiple modified text prompts act as a noisy test-time augmentation for the original one. 
We propose an alternative evaluation scenario to decide if a performance boost of LLM-generated descriptions is caused by such a noise augmentation effect or rather by genuine description semantics. The proposed scenario avoids noisy test-time augmentation and ensures that genuine, distinctive descriptions cause the performance boost. Furthermore, we propose a training-free method for selecting discriminative descriptions that work independently of classname-ensembling effects. Our approach identifies descriptions that effectively differentiate classes within a local CLIP label neighborhood, improving classification accuracy across seven datasets. Additionally, we provide insights into the explainability of description-based image classification with VLMs.
\end{abstract}

\begin{links}
    \link{Code}{https://github.com/CompVis/DisCLIP}
\end{links}

\begin{figure*}[h]
\begin{center}
\includegraphics[width=\linewidth]{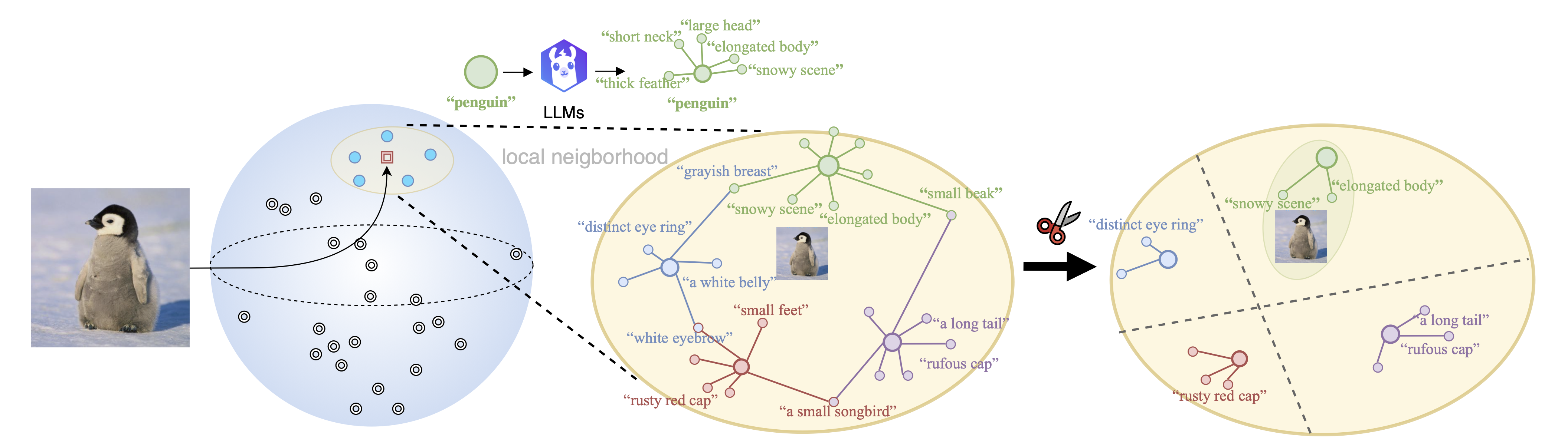}
\end{center}
\caption{Are the extra semantics provided by LLM truly useful? Our method first identifies candidate labels using only the class name. We then filter out descriptions that may seem logical but do not differentiate the group, \eg ambiguous, overly generic, or noisy descriptions. This refinement ensures that the remaining descriptions provide distinctive vision-language cues within the local candidate neighborhood, offering more specificity than the class name alone can capture.
}
\label{fig:teaser}
\end{figure*}

\section{Introduction}

Human visual recognition is closely related to verbal reasoning, as it often relies on the ability to express visual information in words~\cite{zhao-etal-2022-explainable, shtedritski2023does}. However, a neural network usually does not exhibit this property, making its explainability a significant concern for the machine learning community. Some studies~\cite{zhang2024vision,hakimov-schlangen-2023-images} aim to connect visual cues and textual descriptions, but these usually require extensive human subject analysis and highly specific datasets with annotations ~\cite{young2014image}, which are expansive to obtain~\cite{lin2014microsoft}.

Vision-Language Models~\cite{clip,jia2021ALIGN} tackle this issue by training neural networks to link images and their textual descriptions within a shared embedding space. This enhances the correlation between visual and textual details.
VLMs can be applied to zero-shot image classification by passing an image through the VLM’s image encoder and prompting the text encoder with hand-crafted inputs like \textit{``a photo of a [classname]''}~\cite{clip}. Recent works~\cite{dclip,chiquier2025evolving} extends this approach by incorporating additional descriptions generated by Large Language Models (LLMs) for each class name. LLMs like GPT-3~\cite{chatgpt,chatgpt-2} or Llama~\cite{touvron2023llama}, trained on extensive text corpora, are intended to provide richer semantics, enhancing VLM classification.

However, \textit{ Does VLM classification truly benefit from LLM-generated description semantics?} This work explores this core question, as LLM-generated descriptions present several challenges. For example, descriptions can overlap for similar classes — such as parrots and sparrows — both described as having feathers, which is not a distinguishing feature. Moreover, while supplying the model with as many LLM-generated descriptions as possible may seem advantageous, it results in excessively lengthy collections. This complicates understanding the contribution of each description to the final decision.

Another problem is the structured noise ensembling phenomenon~\cite{waffleclip}: LLM-generated descriptions can be replaced with high-level concepts and random characters (such as ``\textit{Baklava}'', ``\textit{a food that is 34mfqr5}'') while still improving the model performance. These slightly modified duplications act as test-time augmentation for the original prompt, resulting in an averaged robust output. This raises the question whether the improvement is due to additional semantics of the LLM-generated descriptions or to the ensemble effect of the noise augmentation.

Given these challenges, a proposed model should meet three criteria: 1) As humans who describe with a limited set of descriptions, the model should also operate with a manageable number of text descriptions. 2) These descriptions should be semantically meaningful. 3) The model should be resilient to noise ensembling. To address the issue of noise ensembling, we constrain the model to use only textual descriptions that do not contain the classname, \ie classname-free descriptions.

Additionally, we employ a training-free algorithm that processes textual description embeddings within the neighborhood of the queried image embedding. This approach narrows the focus to descriptions relevant to distinguishing between the specific subset of ambiguous classes, reducing the information to process and targeting a more manageable problem rather than attempting to differentiate across all classes.  Our method passes the image through the CLIP image encoder, identifies a set of ambiguous class names representing possible candidates, and then applies a straightforward procedure to determine the most distinctive descriptions for these candidates, as shown in \Cref{fig:teaser}. 

Moreover, our method uses the text embedding of the classname only once and subsequently leverages classname-free LLM-generated descriptions. Therefore, we ensure that performance gains are not due to noisy augmentations of the classnames but rather to a semantically meaningful enrichment. In summary, our contribution is threefold: 

\begin{itemize}

\item[-]We propose an alternative evaluation scenario for VLM classification tasks to assess whether performance gains stem from genuine semantic understanding rather than an ensemble effect, which is difficult to discern under conventional setups (\Cref{tab:class_name_less} and \Cref{fig:overall_performances}).

\item[-]Using this alternative setup, we introduce a training-free approach (\Cref{sec: our task}) that narrows the focus to a small neighborhood and selects precise, semantically meaningful, and distinguishing class descriptions to improve the VLM classification performance (illustrated in \Cref{fig:teaser}).

\item[-]Our method achieves improved performance compared to related approaches in two different setups, offering insight into the explainability of fine-grained image classification with VLMs (in \Cref{sec:main_results}).

\end{itemize}


\section{Related Works}

\noindent\paragraph{Vision-language models for classification}

Vision-language models such as~\cite{clip} can be used for classification. Notable training-free approaches that build on top of this include DCLIP~\cite{dclip} and CuPL~\cite{platypus}, where class name texts are augmented with knowledge contained in LLMs to leverage seemingly discriminating characteristics to achieve performance boosts. As \citet{waffleclip} demonstrated, similar effects could be obtained by augmenting the class name with text noise and high-level concepts, raising concerns that many performance gains from DCLIP~\cite{dclip} were not due to additional semantics but rather to introduced noise. FuDD~\cite{fudd} introduced contrastive zero-shot prompting to obtain a more diverse set of text prompts. The disadvantage of these approaches is that they rely on ensembling the description extended class name multiple times to achieve significant gains, making it difficult to separate additional semantics from random augmentations of the class name.

Notable approaches that train in the CLIP embedding space include \citet{concise_and_descriptive}, where nearest neighbors from a pool of text embeddings replace linear weights of a learned dictionary, and LaBo~\cite{labo}, which trained a linear classifier on a wide and global bottleneck of language activations selected for diversity and coverage. \citet{coop} performed non-explainable tuning of text prompt embeddings to optimize classification. In contrast, \citet{zang2024pretrained} trained the last layer of image and text encoders over a concept bottleneck to discover explainable concepts. \citet{feng2023text} trained a sparse logistic regression over a matrix of image-language activations, with the training signal also used to train the image encoder.

In contrast to the methods outlined above, our approach delivers humanly understandable (thus explainable), semantically meaningful, disjoint, and distinguishing language descriptions in text space through a training-free method. It boosts VLMs classification accuracy while providing higher explainability. We will further discuss the better explainability in \Cref{sec:interpretability}.

\noindent\paragraph{Test time (noise) augmentation} Data augmentation involves increasing the diversity of training examples without explicitly collecting new data. It can also be employed at test time to enhance robustness~\cite{cohen2019certified} and improve accuracy~\cite{szegedy2015going, jin2018deep}. Notably, simply adding noise to the input string at different levels ~\cite{kobayashi2018contextual, csahin2022augment, belinkov2018synthetic} or their textual embeddings~\cite{sun2020mixup, chen2020mixtext, hao2023mixgen}, can achieve similar effects on both performance and robustness across various tasks and domains~\cite{feng2021survey}.

Test Time Augmentation TTA introduces an \textit{ensemble} of predictions from several transformed or distorted versions of a given test input to obtain a ``smoothed'' prediction. For example, one could average the predictions from various modified versions of a given string, ensuring that the final prediction is robust to any single unfavorable version~\cite{waffleclip, dclip}. On the other hand, \citet{fudd} used up to hundreds of thousands of descriptions per class, achieving significant improvements in classification accuracy with VLM. However, it is challenging to determine if the performance gains result from the vast ensemble or true information, hence hindering explainability. 


\section{Method}

First, we introduce the conventional task formulation for image classification using Vision-Language Models. We then present our unique approach to this task to enable explainability. Finally, we propose a specific solution to enhance the results further.

\subsection{Background}
\label{sec:default_task}
\subsubsection{VLM for Visual Classification}
The process of image classification by Vision-Language Models (VLMs) occurs as follows: Given an image $x$ and a set of class labels $\mathcal{C}$, one classifies the image $x$ by retrieving the class label $\Tilde{c}$ with the highest vision-language score:

\begin{align}
   \Tilde{c}=\mathop{\argmax}_{c \in \mathcal{C}} s(c,x),
\end{align}
where the vision-language scores $s(c,x)$ use a function $\phi(\cdot, \cdot)$, to calculate similarity scores for image-text embedding pairs. A typical instance of $\phi(\cdot, \cdot)$ is the usual cosine similarity. Traditionally, a vision-language score is obtained in the following way: using the image-text embedding function $e$  of the VLM and given a text representation (a string containing the class name) $t_c$ of class label $c$:
\begin{align}
    s(c,x) = \phi(e(t_c),e(x)),
\end{align}
where $e(\cdot)$ is the image or language embedding. 
Another way to obtain vision-language scores is via ensembling. 
The motivation for ensembling can be derived from how a human describes an object. For example, when describing an apple, we can describe it as a \textit{``green stuff''}, \textit{``a round object''}, or  \textit{``fruit of the same size as an orange''}.

In this case, there is no single text representation $t_c$ for a class $c$ but a set of language representations $\mathcal{D}(c)$ where the ensembling happens over the elements of $\mathcal{D}(c)$:
\begin{align}
    s(c,x) = \frac{1}{|\mathcal{D}(c)|} \sum_{d \in \mathcal{D}(c)} \phi(e(d), e(x)),
\label{eq:conventional_ensembling}
\end{align}
In the course of this work and w.r.t. to textual descriptions, we call a set $\mathcal{D}(c)$ a description assignment. Furthermore, an LLM "assigns" descriptions by generating them when prompted for a particular class $c$, yielding the assignments $D(c)$.
\label{subsec:classname-containing descriptions}
The elements of $\mathcal{D}(c)$ can be pure text augmentations, \eg \textit{``an image of \texttt{[cls]}''}, \textit{``a photo of \texttt{[cls]}''}, or can contain LLM-generated text descriptions and high-level-concepts, \eg \textit{``an image of \texttt{[cls]}, a type of \texttt{[LLM-generated category]}, with \texttt{[LLM-generated descriptions]}''}. 
Most of the approaches~\cite{dclip,platypus,fudd,waffleclip} that use ensembling as in \Cref{eq:conventional_ensembling} with LLM-generated contents always include the class name token \texttt{[cls]} for $\forall d \in \mathcal{D}(c)$, which we denoted as \textit{classname-included descriptions}.

\begin{figure*}[h]
\centering
\includegraphics[width=0.95\linewidth]{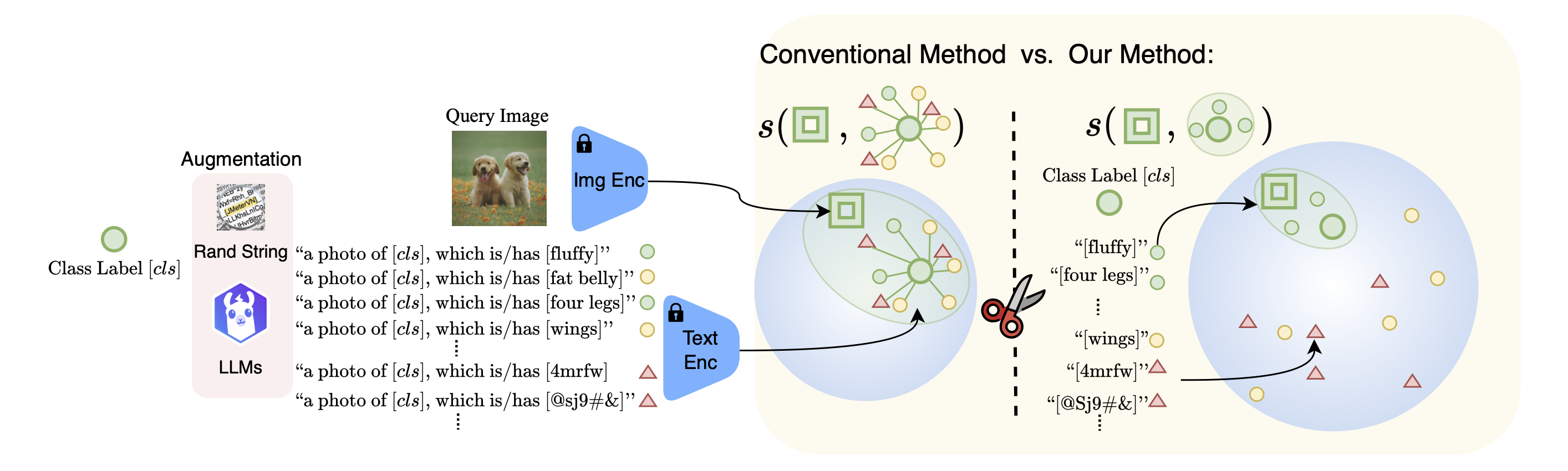}
\caption{
In the conventional setup (\textit{left}), using CLIP with LLM-assigned class descriptions or even random strings can sometimes result in performance gains due to the added semantics or the smoothing ensemble effect. However, when the classname is removed, \ie under the proposed classname-free setup (\textit{right}), these descriptions will fail to perform well, as only meaningful descriptions \wrt the class are useful. In contrast, random strings or non-informative descriptions bring no gain. 
} 
\label{fig:setup_compare}
\end{figure*}

\subsection{Our Approach}
\label{sec: our task}

\subsubsection{Classname-free descriptions}
\label{subsec:classname-free descriptions}
In the conventional setup~\cite{dclip,platypus,fudd}, performance gains may result from the noise augmentation of the class name text \texttt{[cls]} embedding through its various combinations with \texttt{[LLM-generated category]}, \texttt{[LLM-generated descriptions]}, and even random strings. While random strings should not contribute extra semantics and are likely embedded far away from \texttt{[cls]}, this can sometimes apply to LLM-assigned text due to the vocabulary discrepancy between VLMs and LLMs. Another cause may also be that the images do not exhibit the described property. Despite this, such combinations can still perform well, although the assigned descriptions are not semantically correct. 

To better investigate whether the improved performance stems from semantic enrichment or the ensemble effect, we propose an approach where, out of all elements in $\mathcal{D}(c)$, exactly one element should contain the class name $c$. The remaining elements must contain textual descriptions without the class name. This set of descriptions then becomes:

\begin{align}
    \mathcal{D}(c) = \{d^{c+},d^{c-}_{0},...,d^{c-}_{m}\},
\label{eq:our_D(c)}
\end{align}

where $d^{c+}$ denotes that the description contains the class name, while $d^{c-}$ denotes that it does not.
 A typical $\mathcal{D}(c)$ can therefore be the following: \textit{\{``An image of \underline{apple pie}.'', ``crispy brown crust'', ``graham cracker crust''\}}.
Whereas in the conventional setup, the \texttt{[cls]} would be the following \textit{\{``An image of \underline{apple pie}.'', ``An image of \underline{apple pie} with crispy brown crust'', ``An image of \underline{apple pie} with graham cracker crust''\}}. The comparison of different setups is shown in~\Cref{fig:setup_compare}.

\subsubsection{Different weight for \texttt{cls}}
\label{subsec:weights}
As discussed previously, the language-ensembling VLM method is evaluated under the conventional scenario by averaging the aggregated class-specific similarities between the images and class-specific descriptions. 

However, in our classname-free setup, it is unclear if plain averaging across the obtained classname-free descriptions and the single \texttt{cls} is appropriate. This is because the classname is probably the most important text representation of the class, whereas the classname-free descriptions rather have a supporting, distinguishing character.
To address this challenge in our evaluation, a weighting factor $w_{cls} \in \mathbb{R}^+$ gets introduced to the vision-language ensemble:
\begin{align}
    s(c,x) = \frac{1}{|\mathcal{D}(c)|} \sum_{d \in \mathcal{D}(c)} w(d) \cdot \phi(d, x)
\label{eq:our_ensembling}
\end{align}
with
\[
w(d) =
\begin{cases}
    w_{cls} & \text{if } d = d_{cls}, \\
    \frac{1}{|\mathcal{D}(c)|-1} & \text{if } d \in \mathcal{D}(c) \setminus \{d_{cls}\}.
\end{cases}
\]
Weights of the classname-free descriptions are normalized to one to have the same relative weightings between classes with different amounts of assigned descriptions. Nonetheless, the challenge remains how to find class-specific, classname-free descriptions that actually improve the classification accuracy. This we shall discuss next.

\begin{algorithm*}[th]
\caption{Inference: Obtain distinctive language descriptions with feedback from VLM space.
}
\label{alg:pseudocode}
    \begin{algorithmic}[1]
        \Require{$x_i$ - Query image to be evaluated \\%
        $\mathcal{P}$ - global description pool obtained from previous stage \\%
        $\mathcal{I}$ - probing image embeddings containing few $n$ samples $\forall c \in \mathcal{C}$ in training split\\%
        $\mathcal{A}_i$ - a set containing $k$ preliminary labels using standard CLIP retrieval with only $cls$\\%
        $\Phi_m$ - selection heuristic to get $m$ descriptions for $\mathcal{A}_i$ from the pool}

        \Ensure{output a set of distinctive language descriptions $\mathcal{D}_i \in \mathbb{N}^{k \times m}$}
            
        \State $\emph{S}\leftarrow \text{matmul}(\mathcal{I},\mathcal{P})\text{.reshape}(n,|\mathcal{C}|,|\mathcal{P}|).\text{mean}(\text{dim}=0) \in \mathbb{R}^{|\mathcal{C}| \times |\mathcal{P}|}$ \Comment{Look-up similarity matrix}
        
        \State $\mathcal{D}_i \leftarrow \{\}$
            \For{each element $a \in \mathcal{A}_i$}

                \State $\emph{S}^{+}_i \leftarrow [S[a,:] - S[\mathcal{A}_i\setminus a,:]]^{+}$  \Comment{Select the positive subset}

                \State $\mathcal{D}_{i,a} \leftarrow \Phi_m(\emph{S}^{+}_i) \in \mathbb{N}^{m}$ \Comment{Extract $m$ descriptions that distinguish $a$ from the other $\mathcal{A}_i$}
                \State $\mathcal{D}_i \leftarrow \mathcal{D}_i \cup \mathcal {D}_{i,a}$ \Comment{Descriptions to differentiate $x_i$ from the $k$ preliminary labels.}
            \EndFor
            \State $s(\mathcal{D}_i,x_i)$ \Comment{Compute similarity within the local neighborhood}

    \end{algorithmic}
\end{algorithm*}

\begin{figure*}[h]
\centering
    \begin{subfigure}{0.27\textwidth}
        \includegraphics[width=\linewidth]{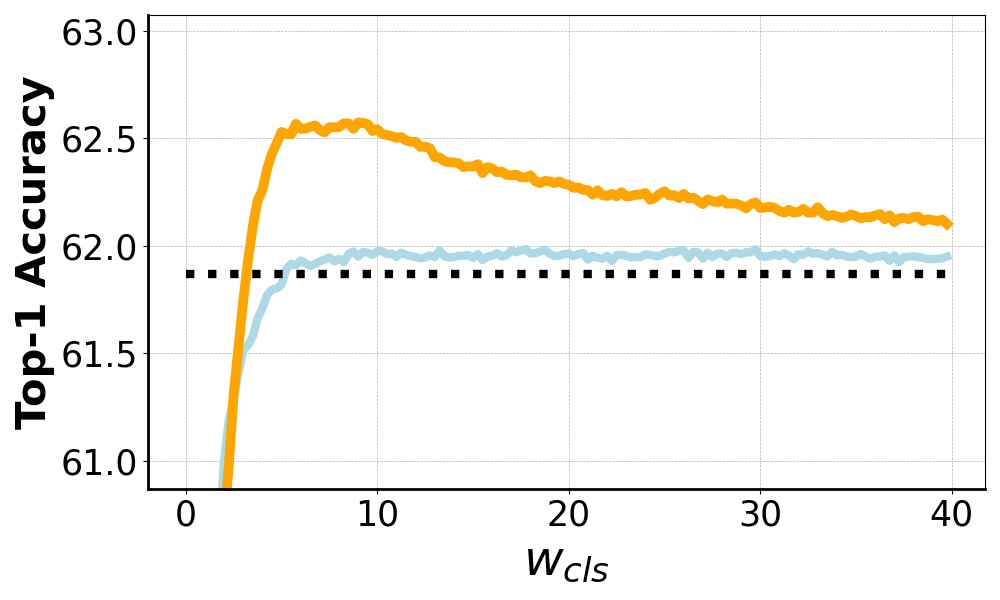}
        \caption{\textbf{ImageNet}}
    \end{subfigure}
    \begin{subfigure}{0.27\textwidth}
        \includegraphics[width=\linewidth]{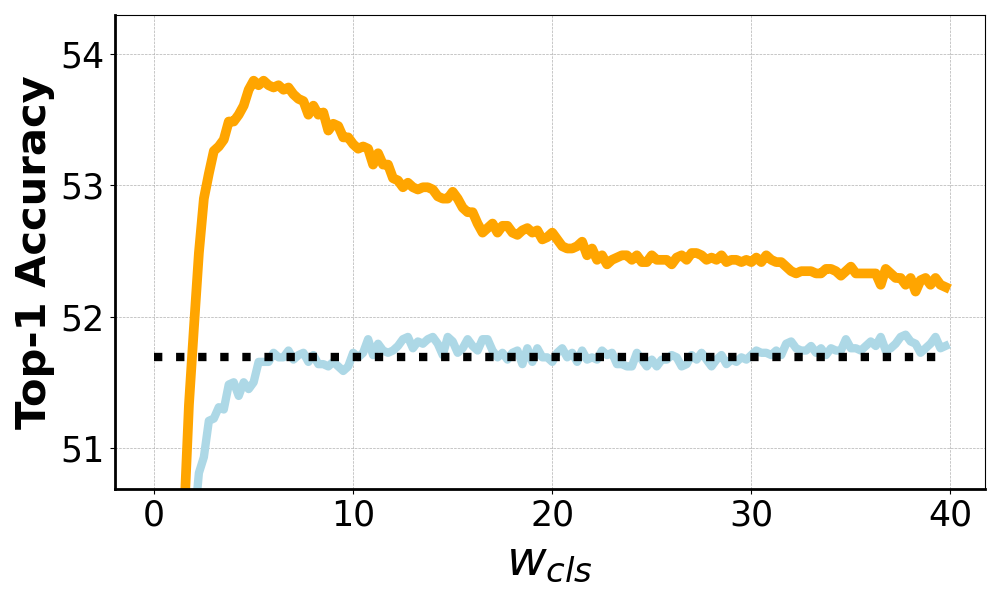}
        \caption{\textbf{CUB200}}
    \end{subfigure}
    \begin{subfigure}{0.27\textwidth}
        \includegraphics[width=\linewidth]{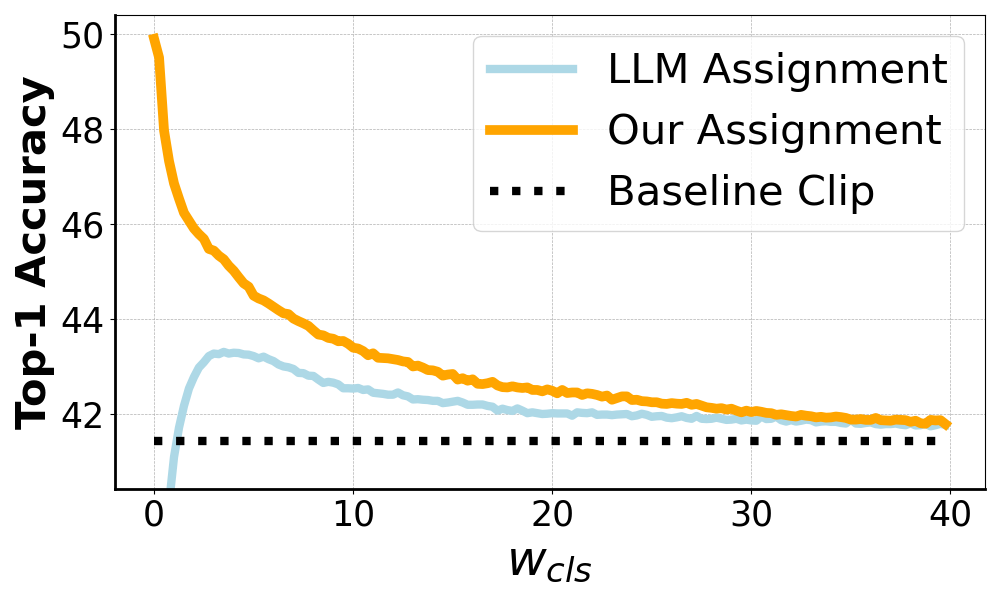}
        \caption{\textbf{EuroSAT}}
        \label{fig:cub}
    \end{subfigure}
    \begin{subfigure}{0.27\textwidth}
        \includegraphics[width=\linewidth]{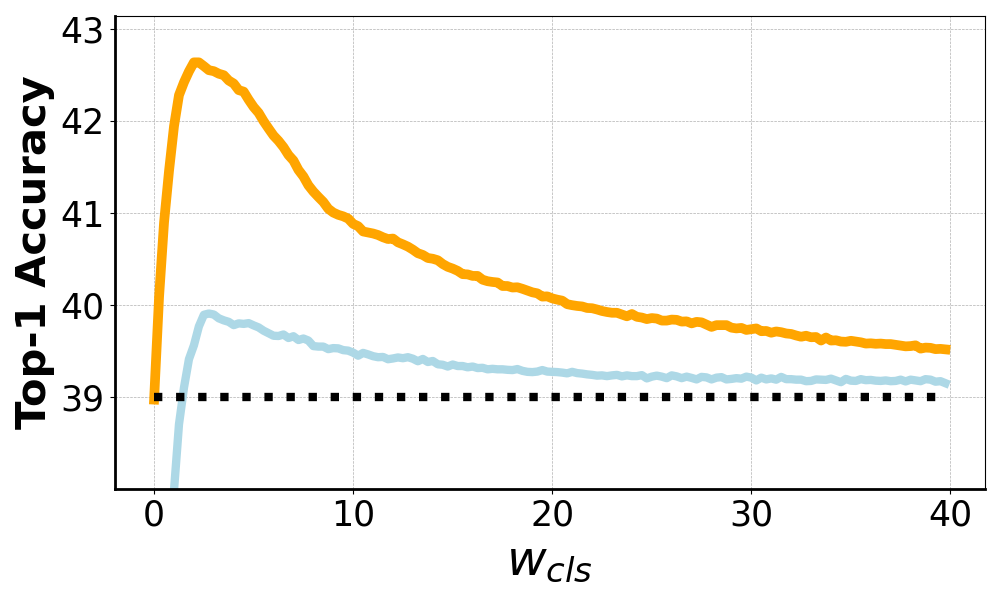}
        \caption{\textbf{Places365}}
    \end{subfigure}
        \begin{subfigure}{0.27\textwidth}
        \includegraphics[width=\linewidth]{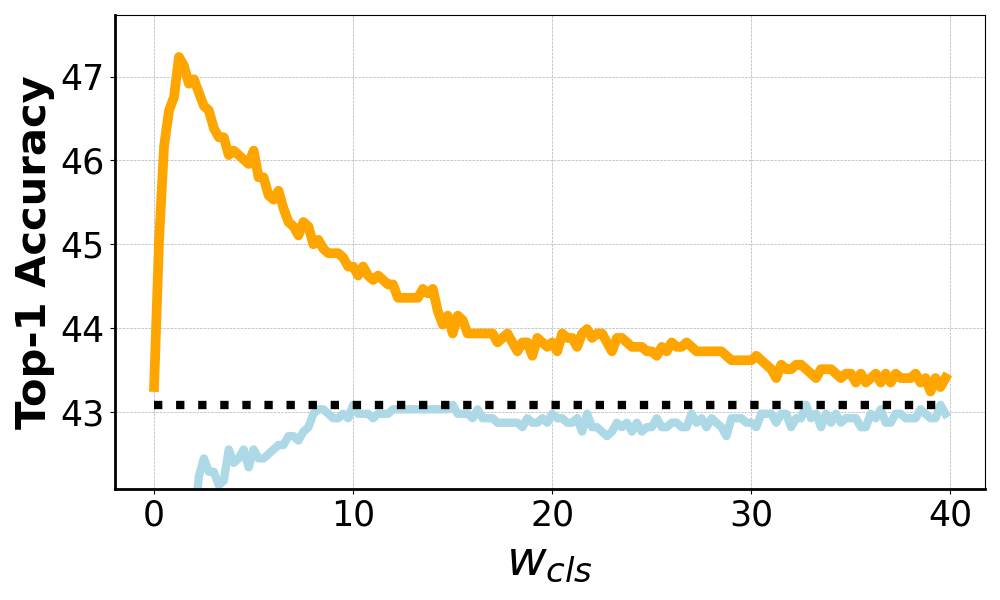}
        \caption{\textbf{DTD}}
    \end{subfigure}
    \begin{subfigure}{0.27\textwidth}
        \includegraphics[width=\linewidth]{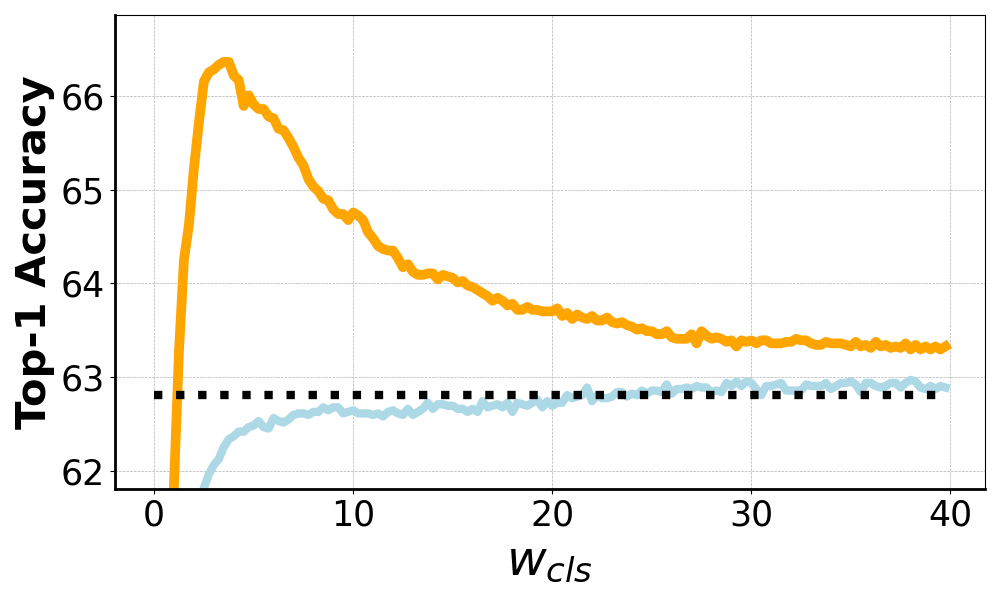}
        \caption{\textbf{\textbf{Flowers102}}}
    \end{subfigure}    
    \caption{Overall Performance of all datasets in classname-free setup. For descriptions assigned by our method and an LLM, $w_{cls}$ assesses the influence of class labels on the performance across different datasets. For a detailed discussion, see \Cref{sec:main_results}.
    } 

\label{fig:overall_performances}
\end{figure*}

\subsubsection{Selection of descriptions}
\label{sec:vlm_feedback}
Our method works in a local candidate neighborhood: Given a test image $x_i$, one retrieves its top-$k$ predictions based solely on text embeddings of texts such as \textit{``a photo of \texttt{[cls]}''}. These preliminary candidate labels constitute the image's local label neighborhood $\mathcal{A}(x_i)= \{\emph{a}_0, a_1, \ldots, a_k\}$, in which more fine-grained descriptions can offer further distinctiveness.

The image-language similarities of class descriptions can correlate positively or negatively with ambiguous candidate classnames of a test image. Ideally, one wants to find descriptions that only correlate positively with one of the ambiguous classnames and negatively with all the others - hence providing a distinctive and explainable language representation. Consequently, the assignment of classname-free descriptions of classes denoted as $\mathcal{D}(c)$ can significantly influence the final classification result. For example, an albatross might be best distinguished from a penguin by \textit{``sailing through the air''} while it might not be well told apart by \textit{``is a seabird''} since both classes share this feature. Furthermore, this connection must also be well represented in the VLM embedding space.

\Cref{alg:pseudocode} depicts our proposed procedure to find such descriptions. Having available $n$ reference image samples per class $c$ and a global, classname-free description pool $\mathcal{P}$ to select from, the goal is to find a set of descriptions $\mathcal{D}(c) \subset \mathcal{P}$ with $|\mathcal{D}(c)|=m$ that distinguishes each class $a \in \mathcal{A}(x_i)$ from its most ambiguous classes $a' \in \mathcal{A}(x_i) \setminus a$, \ie the small neighborhood of classes around the given images, as depicted in the left part of \Cref{fig:teaser}. For that, one utilizes a lookup matrix $S$ containing classwise averaged image-description similarities to obtain feedback from the VLM embedding space. The criterion for assigning descriptions $D(a)$ is a score that is positive if a description activates on average higher for $c=a$ than for all $a' \in \mathcal{A}(x_i) \setminus a$, \cf line 4 of \Cref{alg:pseudocode}. This yields $\emph{S}^{+}_{i}$, a positive subset of the lookup similarity matrix $S$.

As $|\emph{S}^{+}|>m$ and one wants to extract the most distinctive descriptions from it, the selection heuristic $\Phi$ gets applied to $\emph{S}^{+}$. It selects top-$m$ scoring descriptions via:
\begin{align}
    \texttt{top-$m$}(\emph{mean}(\emph{S}^{+},dim=0)), 
\end{align}
 \ie, those $m$ descriptions whose averaged image-description similarity differences to $c$ are, on average, maximally large. In other words, these descriptions activate highly for $a$ but not highly for any $a' \in A \setminus a$ on average. Because these $m$ descriptions are selected without a prepended class name $\emph{cls}$, they can serve as classname-free language representations of class $c$.

The selected descriptions can then be used as described in \Cref{subsec:classname-free descriptions} or \Cref{subsec:classname-containing descriptions} for inference. In both cases, classification happens via an ensembling (classname-containing and classname-free) of image-language similarities, as introduced in DCLIP. Applying $\argmax$ over the ensembled, description-enriched image-language similarities of the candidate set $\mathcal{A}(x_i)$ yields the final classification decision of the image.

\subsubsection{Better explainability}
\label{sec:interpretability}
Our proposed method achieves \textit{better explainability} by offering these four characteristics:
\begin{enumerate}
    \item \textbf{The original CLIP encoders for text and image are retained,} rather than fine-tuned to represent a different embedding space, as seen in some works~\cite{zang2024pretrained,feng2023text}. Hence, our approach preserves the general validity of the CLIP embeddings.
    
    \item \textbf{The number of resulting textual descriptions for a single class is kept within a reasonable limit,} similar to the approach in the seminal work~\cite{dclip}. This helps minimize the potential for noise augmentation, unlike methods that generate hundreds and thousands of descriptions~\cite{fudd}.
    
    \item \textbf{The overlap between concepts across various classes is minimized,} in comparison to methods with global concept bottlenecks~\cite{labo,concise_and_descriptive,zang2024pretrained}. Sparse overlapping ensures clearer distinctions between classes.
    
    \item \textbf{We do not use continuous weights over resulting textual descriptions,} as done in ~\citet{labo,concise_and_descriptive,zang2024pretrained}. Long vectors of continuous weights can be less interpretable compared to clear, discrete indicators of whether a concept is present. Hence, our method offers improved clarity and explainability.
\end{enumerate}

\begin{table*}[ht]
    \centering
    \setlength{\tabcolsep}{1mm}
    \footnotesize
    \begin{tabular}{ll|c|cccccccc}
    \toprule
    \textbf{Source of $\mathcal{P}$} & \textbf{Description Assignment} & \textbf{Max \#desc. $\downarrow$ } & \textbf{ImageNet}& \textbf{ImageNetV2} & \textbf{CUB200} & \textbf{EuroSAT} & \textbf{Places365} & \textbf{DTD} & \textbf{Flowers102}\\
    \midrule
    DCLIP & LLM (global eval)&  13  & 61.99 & 55.09 & 51.79 & 43.31 & 39.91  & 43.09 & 62.97 \\
    DCLIP & LLM (local-k eval) & 13  & 61.99 & 55.06 & 51.83 & 43.29 & 39.87& 43.09 & 62.86 \\
    DCLIP & \textit{Ours} &  5 &  \underline{62.57} & \textbf{55.48} & \textbf{53.80}  & \textbf{49.89}  & \underline{42.64}  & \textbf{47.23} & \underline{66.37}\\
    
    \midrule
    Random & \textit{Ours } & 5 & 62.18 & \underline{55.22} & 52.31  &  40.82 & 40.44  & 44.73 & 66.12\\
    
    \midrule
    
    Contrastive  & LLM & 40  & 62.03 &54.88 & 52.24  & 46.97  & 40.37  & 44.41 & 62.90\\
    
    Contrastive & \textit{Ours} & 5 & \textbf{62.78} & \textbf{55.48} & \underline{53.45}  & \underline{49.47}  & \textbf{42.65}  & \underline{46.97} & \textbf{67.07}\\
    
    \bottomrule
    \end{tabular}
\caption{Image classification in classname-free setup with different assignments and pools. Our method consistently produces the highest accuracies in this setting. We use the best-performing $w_{cls}$ of the respective assignment to ensure a fair comparison. }
\label{tab:class_name_less}
\end{table*}


\section{Experiment}
This section evaluates our approach on seven widely used benchmark datasets for (fine-grained) visual classification. We compare our approach to state-of-the-art methods and provide qualitative results.

\subsection{Implementation Details and Datasets}
\label{sec:experimental details}

\noindent\textbf{Implementation details.} We use CLIP~\cite{clip} as the base Vision-Language Model (VLM) for our approach. Unless stated otherwise, the backbone for CLIP is the ViT-B/32~\cite{vaswani2017attention, dosovitskiy2020image}. We randomly sample a subset from each dataset's standard training split to obtain the lookup similarity table $\emph{S}$ (details see \Cref{app:hyperparameter_n}). Our empirical tests confirmed that this sampling process does not significantly impact performance. The Large-Language Model (LLM) generated descriptions are sourced directly from DCLIP~\cite{dclip} or generated using the contrastive prompting method with \texttt{gpt-3.5-turbo-1106} and \texttt{Llama-3-70b-chat-hf} via APIs.

\noindent\paragraph{Datasets.} We evaluated our methods on the following standard datasets using the standard protocol (classification accuracy) based on previous works~\cite{dclip, waffleclip}: ImageNet~\cite{imagenet}, ImageNetV2~\cite{recht2019imagenetV2}, CUB200-2011~\cite{cub200-2011} (fine-grained bird classification), EuroSAT~\cite{eurosat} (satellite image recognition), Places365~\cite{places365}, DTD (Textures, \cite{dtd}), and Flowers102 \cite{flowers102}.

\noindent\paragraph{Source of obtaining description pool $\mathcal{P}$.}
These descriptions can be obtained in the following ways: 1) directly from the published descriptions of other works, such as DCLIP ~\cite{dclip} or FUDD ~\cite{fudd}; 2) generated based on the provided procedures and code bases of other works, if descriptions are not available; 3) or created through contrastive prompting, which aims to extract meaningful descriptions by contrasting hard negative samples within a neighborhood.
The motivation is similar to FuDD ~\cite{fudd}, but we use significantly fewer descriptions per class. As this is only an alternative for constructing a description pool and orthogonal to our proposed method, we provide more details in \Cref{app:construction_pool} on the construction of the contrastive pool.

\begin{table*}[th]
\centering
\footnotesize
\begin{tabular}{l|ccccccc}
\toprule
 \textbf{Description Assignment}  & \textbf{ImageNet} & \textbf{ImageNetV2} & \textbf{CUB200} & \textbf{EuroSAT} & \textbf{Places365} & \textbf{DTD} & \textbf{Flowers102} \\
\midrule
 LLM assignments & 11.65 & 10.69& 3.47 & 28.11 & 21.36  & 17.77 & 3.19 \\
random assignments & 0.08  & 0.06&  0.43  & 11.61  & 0.11  & 2.45 & 1.01\\
\midrule
\textit{Ours} & \textbf{50.16} & \textbf{43.98} & \textbf{41.53}  & \textbf{43.24}  & \textbf{36.36}  & \textbf{43.09} & \textbf{51.52}\\
\bottomrule
\end{tabular}

\caption{Performance in classname-free setup with $w_{cls}=0$. Our descriptions are robust and perform well, even if the classname text $D_{cls}$ is weighted by $w_{cls}=0$. LLM assignments give a considerably worse performance in this scenario. Randomly assigned descriptions fail to provide reasonable guidance. Llama3-70B with Contrastive Prompting is used as source pool $\mathcal{P}$. Ambiguous context size $k=3$. For sample sizes $n$ see~\Cref{app:hyperparameter_n}.} 
\label{tab:0cls_weight}
\end{table*}

\begin{table*}[th]
\centering
\footnotesize
\setlength{\tabcolsep}{1mm}
\begin{tabular}{l|l|c|ccccccc}
\toprule
\textbf{Method} &\textbf{Source of $\mathcal{P}$} & \textbf{Max \#desc.} $\downarrow$ & \textbf{ImageNet} & \textbf{ImageNetV2} & \textbf{CUB200} & \textbf{EuroSAT} & \textbf{Places365} & \textbf{DTD} & \textbf{Flowers102}\\
\midrule
CLIP & - & 1 & 61.87 & 54.74 & 51.69 & 40.92 & 39.01 & 43.09 & 62.81 \\

DCLIP & DCLIP & 12 &  62.22 & 54.84 & 52.55 & 47.33 & 40.01  & 41.86 & 62.17 \\

WaffleClip & WaffleClip  & 30 & 63.31 & \underline{55.92} & 52.38  & 44.31  & 40.56  & 43.16 & 66.27\\

FuDD & FuDD & 1842 & \textbf{64.19} & \textbf{56.75}& 54.30  &45.18  & 42.17 &  44.84 & 67.62\\
\midrule
\textit{Ours }& DCLIP &  5  & 61.59 &53.61 & 55.89  & \underline{50.05}  & 42.77  & \textbf{48.83} & 66.99\\
\textit{Ours}& Contrastive & 20  & 63.30 &55.24 & 56.27 & \textbf{58.57} & \textbf{43.65} & \underline{48.09} & \underline{68.61} \\
\textit{Ours}& FUDD &25 & 61.86 & 53.05 & \textbf{56.62} & 48.42 & 42.76   & 48.03  & 68.47\\
\textit{Ours}& Contrastive &50 & \underline{63.51} &55.41 & \underline{56.45} & 44.46 & \underline{43.62}   & 47.66  & \textbf{69.51}\\
\bottomrule
\end{tabular}

\caption{Image classification with classname included in the descriptions. Ambiguous context size $k=3$. For sample sizes $n$ see~\Cref{app:hyperparameter_n}. An ablation of ambiguous context size $k$ can be found in~\Cref{app:k_ablation}.
}
\label{tab:main_convention}
\end{table*}

\subsection{Experimental Results}
\label{sec:main_results}
\noindent\paragraph{Classname-free evaluation.} We evaluate the quality of the classname-free description assignments selected by our method in the classname-free evaluation setup (cf. ~\Cref{subsec:classname-free descriptions}). Examples of selected descriptions can be found in ~\Cref{subsec:app:example_descriptions}.
Performance of our algorithm across $7$ classification benchmarks is shown in~\Cref{fig:overall_performances}, highlighting how varying $w_{cls}$ impacts top-$1$ accuracy. The non-ensembled CLIP baseline performance, independent of $w_{cls}$, is also included for reference. 
Our selected assignments consistently outperform the DCLIP LLM assignments. Notably, for the EuroSAT, Flowers102, CUB200, DTD, and Places datasets, optimal performance occurs when $w_{cls}$ is low ($[0,10]$), emphasizing the importance of classname-free descriptions while exceeding the baseline performance by up to $9\%$ and the LLM performance by up to $8\%$. However, the LLM-assigned descriptions cannot produce performance gains in the classname-free scenario that comes close to our selections. 

Further increasing $w_{cls}$ and thereby weighting the single classname-included description higher reduces accuracy, showing that overly prioritizing the classname diminishes the benefits of our classname-free descriptions.

Interestingly, the smaller gain for ImageNet ($\approx0.5$pp.) also corresponds to a lower bump for low $w_{cls}$ in the plot. This may be due to the noisier backgrounds of this dataset, which hinders the selection of generally valid descriptions.

Quantitative results are shown in ~\Cref{tab:class_name_less}, where we report the peak accuracy for each dataset regardless of $w_{cls}$.
Interestingly, only $5$ selected classname-free descriptions per preliminary class of an image are enough to surpass the performance of the DCLIP LLM assignments. An additional classname-free performance of up to $6.6$ pp. (for the EuroSAT dataset) can be achieved. 

To confirm that our gains are not driven solely by the image-wise top-$k$ neighborhood, we also evaluate DCLIP LLM assignments in the local top-$k$ context, which shows no significant improvement.
This suggests that our approach succeeds by the selection procedure within the top-$k$ neighborhood rather than the search space restriction alone. Importantly, these gains are independent of a specific description pool $\mathcal{P}$ as they also hold for a contrastive prompting pool.

To our knowledge, no prior work has explored a comparable classname-free evaluation setup to determine the true distinctiveness of assigned descriptions $d^{c-}_{0},...,d^{c-}_{m}$ in combination with a classname prompt $d^{c+}$. However, some works use methods like trainable bottleneck classifiers ~\cite{labo,concise_and_descriptive} or trainable embeddings~\cite{zang2024pretrained}, which can be considered "classname-free." Despite this similarity, they are too different to compare against (detailed discussion in \Cref{subsec:app:classname-free_compare}).

\noindent\paragraph{Conventional setup.}
We evaluate our chosen descriptions in a conventional setup where classnames are included in all descriptions, as shown in \Cref{tab:main_convention}. 
Our method performs well on datasets where a higher performance bump is observed with low $w_{cls}$, \ie high relative description weights in \Cref{fig:overall_performances}. This happens when the selected description pool provides generalizable, diverse, and discriminative descriptions for the datasets. Our method outperforms DCLIP assignments in the DCLIP pool and outdoes WaffleCLip and FuDD on datasets CUB200, EuroSAT, Places365, DTD, and Flowers102. This remains true for $4$ of these datasets, even if only $5$ descriptions per class are used. In contrast, WaffleClip~\cite{waffleclip} uses $30$ text prompts per class, and FuDD~\cite{fudd} uses an astonishing number of {1,842} descriptions per class.

On the other hand, for ImageNet and ImageNetV2, we can see a connection between suboptimal conventional performance and a much lower peak relative to baseline CLIP in the classname-free setting - indicating less distinctive power of our assignments. Mixed results in the conventional setup for ImageNet and ImageNetV2 imply it is challenging to find distinctive descriptions - at least within the currently used description pools $\mathcal{P}$. This difficulty may arise because random image contents, \eg, background objects, distort the description assignments.
Our algorithm experiences a performance boost when using a $\mathcal{P}$ obtained through contrastive prompting, offering a richer pool of descriptions.

Overall, the results from the class name-containing scenario suggest that the added semantics of the discovered descriptions enhance the performance—in addition to the class name ensembling used by other methods like WaffleClip, FuDD, and DCLIP.

\noindent\paragraph{Performance in classname-free scenario when $w_{cls}=0$.} We evaluate the performance under a classname-free scenario in~\Cref{tab:0cls_weight} without any guiding classname information. In this case, random assignments don't achieve any reasonable classification accuracy; LLM-assigned descriptions provide minor guidance. With our selected descriptions, however, we have achieved decent performance across all datasets - significantly surpassing the LLM assignments. This further supports the idea that descriptions assigned by an LLM are not distinctive enough. Instead, feedback from the embedding space is needed for distinctive assignments. Higher distinctiveness also shows in~\Cref{subsubsec:max_vs_mean} where classname ensembling is prohibited via a maxing-aggregation.


\section{Conclusion}
This study demonstrates that \textit{VLM Classification performance indeed benefits from LLM description semantics - if the descriptions are correctly selected}. To achieve this, we introduce a training-free method that assigns semantically meaningful descriptions based on feedback from the VLM embedding space. Our results indicate that these descriptions possess inherent discriminative power, as evidenced by evaluations conducted without classname ensembling in our proposed setup. Furthermore, incorporating these description assignments enhances performance in image classification tasks, both with and without classname ensembling. Additionally, our evaluation framework effectively distinguishes performance improvements arising from genuine semantic understanding from those resulting from ensemble effects. We hope that our findings will inspire future research on VLMs and contribute to the development of models with enhanced explainability.

\section*{Acknowledgements}
This project has been supported by the German Federal Ministry for Economic Affairs and Climate Action within the project ``NXT GEN AI METHODS – Generative Methoden für Perzeption, Prädiktion und Planung'', the German Research Foundation (DFG) project 421703927, Bayer AG, and the bidt project KLIMA-MEMES. The authors gratefully acknowledge the Gauss Center for Supercomputing for providing compute through the NIC on JUWELS at JSC and the HPC resources supplied by the Erlangen National High Performance Computing Center (NHR@FAU funded by DFG).


\bibliography{aaai25}


\clearpage

\appendix
\section{Appendix}
\label{sec:appendix}

\subsection{Limitations and Ethical Considerations}
Currently, most of the methods in this domain, including ours, work with a fixed pool of descriptions $\mathcal{P}$ from LLMs. Although our method finds precise and meaningful descriptions for better performance, it would be interesting to give such selection feedback to LLMs and let them refine the $\mathcal{P}$ in an agentic fashion. We are not currently aware of any ethical considerations.

\subsection{Different choices of size  \texorpdfstring{$k$}{k} and  \texorpdfstring{$n$}{n}}
\label{app:hyper}
~\Cref{tab:hyper} ablates the number $k$ of preliminary labels per test image and $n$, the number of images per class available in the selection set for $\emph{S}$. Results were obtained in the classname-included evaluation setup on the CUB200 Dataset with $5$ selected descriptions from the DClip description pool. Increasing the number of selection samples in $S$ leads to increasing performances. For the CUB dataset, the optimal $k$ value is empirically found to be $k=3$ as it has a right trade-off between a diverse set of candidates that is still small enough to find distinctive descriptions. This shows that even in a few-shot regime with only $5$ selected descriptions per candidate class baselines such as ~\citet{fudd} and ~\citet{dclip} can be exceeded (cf. ~\Cref{tab:main_convention} in the main text).

\begin{table}[h]
\centering
\begin{tabular}{l|cccc}
\toprule
\textbf{$n$ / $k$} & \textbf{2} &  \textbf{3} &  \textbf{4}\\
\midrule\
5  & 53.33  & 54.00  & 54.04 \\

10  & 54.38 & 55.20  & 54.31  \\

15 & 54.85 &  55.63  & 55.51  \\

20 & 55.28 &  \textbf{55.82} & 55.75 \\

25 & 55.70 &  55.59 & 55.44 \\

\bottomrule
\end{tabular}
\caption{Hyper-params comparison on CUB200 dataset for ViT-B/32 backbone. $k$ defines the number of preliminary labels per test image to consider, and $n$ denotes the number of images per class available in the selection set for $\emph{S}$.
}
\label{tab:hyper}
\end{table}

\subsection{Ablation for description assignments with \texorpdfstring{$\mathcal{P}$}{P} from DCLIP.} 
An intriguing experiment investigates what happens if descriptions get assigned randomly to classes. ~\Cref{tab:assigment} compares LLM assignments to random assignments and the assignments of our method in the classname-free setup. The assignments are evaluated both on a global scale, without restricting the classification to local candidates, and a local top-$k$ candidate neighborhood (cf. ~\Cref{alg:pseudocode}). As can be seen, the LLM assignments provide some guidance but not reliably so. They perform similarly to random assignments, and even for 2 datasets random descriptions can surpass the LLM assignments. The effect of the local candidate evaluation for random and LLM assignments is restricted (rows 2 and 4),  as their description assignments are not adjusted to the local candidate set. On the other hand, our assignments constantly offer the best results.

\begin{table*}[h]
\centering
\footnotesize
    \begin{tabular}{l|c|cccccccc}
    \toprule
    \textbf{Description Assignment} & \textbf{Max \#desc.} $\downarrow$ & \textbf{ImageNet} & \textbf{ImageNetV2} & \textbf{CUB200} & \textbf{EuroSAT} & \textbf{Places365} & \textbf{DTD} & \textbf{Flowers102}\\
    \midrule
    LLM &  13  & \underline{61.99} & \underline{55.09} & 51.79 & 43.31 & \underline{39.91}  & 43.09 & \underline{62.97} \\
    LLM (local-k) & 13  & \underline{61.99} & 55.06 & 51.83 & \underline{43.29} & 39.87& 43.09 & 62.86 \\
    random &  13  & 61.88 & 54.86 & \underline{52.00}  & 41.46  & 38.98  & \underline{43.24} & 62.81 \\
    random (local-k) &  13  & 61.91 & 54.86 & 51.97  & 41.41  & 39.02   & 43.30 & 62.82 \\
    \midrule
    \textit{Ours} &  5 & \textbf{62.57} & \textbf{55.48} & \textbf{53.80}  & \textbf{49.89}  & \textbf{42.64}  & \textbf{47.23} & \textbf{66.37}\\
    \bottomrule
    \end{tabular}
\caption{Ablation for assignment with $\mathcal{P}$ from DCLIP. Evaluation under the classname-free setup.}
\label{tab:assigment}
\end{table*}

\subsection{Results for ViT-L/14 CLIP Backbone}
We also show evaluation results on the ViT-L/14 CLIP backbone, as it is commonly used for an additional comparison. One can see a consistent pattern as we showed in the main paper for the ViT-B/32. See~\Cref{tab:class_name_less_vit_l_14} and~\Cref{tab:main_vit_l14}.

\begin{table*}
\centering\resizebox{\linewidth}{!}{
\begin{tabular}{l|l|c|ccccccccc}
\toprule
\textbf{Method} &\textbf{Source of $\mathcal{P}$} & \textbf{Max \#des} & \textbf{ImageNet} & \textbf{ImageNetV2} &\textbf{CUB200} & \textbf{EuroSAT} & \textbf{Places365} & \textbf{Food101} & \textbf{Oxford Pets} & \textbf{DTD} & \textbf{Flowers102}\\
\midrule
CLIP & - & 1 & 73.43 & 67.86 & 62.22 & 55.02 & 40.62 & 90.75 & 93.24 & 52.45 & 74.81 \\

DCLIP & DCLIP & 12 &  74.47 & 68.74 & 63.39 & 57.97 & 41.66 & 89.94 & 93.51 & 54.73 & 76.83 \\
WaffleClip & WaffleClip  & 30 & 75.30 & 69.48 & 64.18  & 61.17  & 42.26  & 93.31  & 91.98  & 53.94 & -\\

FuDD ($k=|C|$) & FuDD & 1842 & \textbf{77.00} & \textbf{71.05} & 66.03  & 60.64  & 44.09  & \textbf{94.27}  & 93.51  &  57.23 & 79.67\\
\midrule
\textit{Ours }& DCLIP &  5  & 74.37 & 68.15 &67.79  & 65.05  & 44.16  & 90.11  & 93.05  & 58.83 & 79.69\\
\textit{Ours}& DCLIP & 50  & 75.62 & 69.70 & 68.55 & \textbf{63.71} & 45.00 & 91.54 & \textbf{93.59} & \textbf{59.57} & \textbf{81.05} \\
\textit{Ours}& Contr. Pr. &50 & 76.04 & 69.79 &\textbf{69.07} & 62.77 & \textbf{45.20}  & 91.34  & 93.43  & \textbf{59.57}  & 80.96\\

\bottomrule
\end{tabular}
}
\caption{Image classification with classname-containing descriptions for the ViT-L-14 backbone. The same parameters as in~\Cref{tab:main_convention} were used.
}
\label{tab:main_vit_l14}
\end{table*}

\begin{table*}
\centering\resizebox{\linewidth}{!}{
\begin{tabular}{ll|c|cccccccccc}
\toprule
\textbf{Source of $\mathcal{P}$} & \textbf{Assignment} & \textbf{Max \#desc. $\downarrow$ } & \textbf{ImageNet} & \textbf{ImageNetV2} &\textbf{CUB200} & \textbf{EuroSAT} & \textbf{Places365} & \textbf{Food101} & \textbf{Oxford Pets} & \textbf{DTD} & \textbf{Flowers102}\\
\midrule
DCLIP  & LLM &  12  & 73.66 & 68.05 &  63.07 & \textbf{56.85} & 41.54 & 90.80 & 93.32 & 53.56 & 74.84 \\

DCLIP & \textit{Ours} &  5 &  \textbf{74.55} & \textbf{68.91} &  \textbf{65.59}  & 56.08  & \textbf{44.21}  & \textbf{90.92}  & \textbf{93.35}  & \textbf{57.23} & \textbf{80.45}\\

\midrule
Contrastive  & LLM & 40  & 73.57 & 68.04 & 62.82  & 55.91  & 41.85  & 90.82  & \textbf{93.73}  & 54.84 & 75.09\\

Contrastive  & random & 40  & 73.07 & 67.96 &  63.38  & \textbf{56.39}  & 40.62  & 90.78  & 93.24  & 52.87 & 74.86\\

Contrastive & \textit{Ours} & 5 & \textbf{74.61} & \textbf{68.70} &  \textbf{66.59}  & 55.68  & \textbf{44.34}  & \textbf{91.02}  & 93.30  & \textbf{56.97} & \textbf{80.45}\\

\bottomrule
\end{tabular}
}
\caption{Image classification in classname-free setup for the ViT-L-14 backbone. Maximum value for the best performing $w_{cls}$ per cell. The same parameters as in~\Cref{tab:class_name_less} were used.}
\label{tab:class_name_less_vit_l_14}
\end{table*}

\subsection{Ablating \texorpdfstring{$k$}{k}}
\label{app:k_ablation}

\Cref{tab:k_eval} ablates parameter $k$, which denotes the number of ambiguous classes considered per test image $x_i$. Different datasets behave differently under a changed $k$. Interestingly, datasets with challenging description behavior - namely Food101, ImageNet and ImageNetV2 - do not show a high sensitivity to a varying $k$. The best and worst performances for these datasets do not differ much, although a tendency for slightly better performances for higher values of $k$ can be observed. The remaining datasets with favorable description behavior are much more sensitive to the chosen values of $k$: CUB200 and EuroSAT perform best for low values of $k$ while DTD, Flowers102, and Places365 perform best for medium to high values of $k$. However, for any chosen value of $k$, LLM assignments are surpassed significantly for these datasets. Similar results are confirmed in~\Cref{tab:k_eval_L_bbn} for the ViT-L/14 backbone where Food101, ImageNet, and ImageNetV2 classification performance reacts insensitively to $k$. Interestingly, for ViT-L/14, high $k$ values benefit the classification performance, whereas for ViT-B/32, low $k$ values yielded the best results.

Besides that, it is remarkable that the language-maxing accuracy denoted in parentheses closely matches or even surpasses\footnote{In case of the ViT-L/14 backbone.} the ensembling accuracy, although it cannot make use of the smoothing ensembling effect that can work with random descriptions (see~\Cref{subsubsec:max_vs_mean} for more information). This points to the distinctive quality of the selected descriptions.

\begin{table*}[]
\centering\resizebox{\linewidth}{!}{
\begin{tabular}{c|ccccccccc}
\toprule
\textbf{ViT-B/32}&&&&&\textbf{Test Dataset}\\
\midrule
\textbf{$k$}  & CUB200 & DTD & EuroSAT & Flowers102 & Food101 & ImageNet & ImageNetV2 & Oxford Pets & Places365 \\
\midrule
2 & \textbf{55.90} (55.01) & 46.17 (45.64) & \textbf{55.39} (\underline{55.57}) & 66.11 (66.97) & 80.34 (\underline{80.59}) & 61.64 (\underline{62.38}) & 54.10 (\underline{54.95}) & 83.57 (84.11) & 42.27 (42.05) \\
3&55.89 (\underline{55.63}) & 48.83 (48.09) & 53.41 (53.10) & 66.99 (68.29) & 80.09 (80.27) & 61.58 (62.00) & 53.61 (54.42) & 82.77 (84.16) & 42.77 (\underline{42.49}) \\
4&55.13 (55.44) & 49.57 (48.78) & 48.63 (51.30) & 67.54 (67.80) & 80.06 (80.28) & 61.32 (61.76) & 53.51 (53.85) & 82.86 (83.76) & 42.90 (42.42) \\
5&55.78 (55.40) & 50.05 (\underline{49.26}) & 49.00 (51.81) & 67.49 (68.12) & 80.11 (80.25) & 61.10 (61.58) & 53.43 (54.06) & 82.15 (83.32) & 42.83 (42.27) \\
6& 55.64 (54.92) & 50.05 (48.14) & 48.19 (53.36) & 68.06 (68.27) & 79.91 (80.10) & 61.25 (61.39) & 53.63 (54.04) & 81.96 (82.99) & 42.82 (42.33) \\
7& 55.28 (55.07) & 49.89 (47.77) & 49.14 (53.23) & 68.79 (68.71) & 80.06 (80.17) & 61.30 (61.46) & 53.61 (54.02) & 82.31 (82.34) & 42.75 (42.24) \\
8& 55.28 (54.64) & 50.74 (47.77) & 51.06 (52.16) & 68.74 (68.30) & 79.99 (80.24) & 61.34 (61.44) & 53.91 (54.02) & 82.31 (82.67) & 42.69 (42.06) \\
9& 54.80 (54.61) & \textbf{50.80} (47.55) & 51.57 (49.33) & 68.84 (68.40) & 80.07 (80.31) & 61.25 (61.44) & 53.79 (53.81) & 82.31 (82.45) & 42.84 (42.21) \\
10&54.61 (54.49) & 50.37 (47.18) & 51.94 (50.26) & 69.13 (68.66) & 80.23 (80.28) & 61.39 (61.37) & 53.74 (53.80) & 82.69 (82.69) & 42.75 (42.09) \\
15 &54.50 (53.56) & 50.05 (47.18) & - & 69.87 (68.73) & 80.49 (80.24) & 61.59 (61.50) & 53.95 (53.78) & 83.59 (83.78) & 42.82 (42.08) \\
20 & 53.85 (52.88) & 49.68 (46.54) & - &  70.14 (68.47) & \textbf{80.55} (79.92) & 61.56 (61.26) & 53.84 (53.61) & \textbf{84.60} (\underline{84.85}) & \textbf{43.07} (42.13) \\
25 & 53.54 (52.59) & 49.68 (47.29) & - & \textbf{70.14} (\underline{69.07}) & 80.52 (79.55) & 61.60 (61.14) & 54.16 (53.51) & 84.00 (84.44) & 42.87 (41.88) \\
30 & 53.30 (52.49) & 49.20 (46.49) & - & 69.90 (69.30) & 80.37 (79.47) & \textbf{61.71} (61.17) & \textbf{54.19} (53.47) & 84.03 (84.30) & 42.85 (41.74) \\
\midrule
LLM Assignments & 52.55 & 41.86 & 47.33 & 62.17 & 79.64 & 62.22 & 54.84 & 84.66 & 40.01 \\
\midrule
Maximum Gain in pp. & \textcolor{green}{3.45} & \textcolor{green}{8.94} & \textcolor{green}{8.24} & \textcolor{green}{7.97} & \textcolor{green}{0.95} & \textcolor{green}{0.16} & \textcolor{green}{0.11}  & \textcolor{green}{0.19} & \textcolor{green}{3.06}  \\
\bottomrule

\end{tabular}
}
\caption{Ablation of $k$ for the selected assignments in the classname-containing setup for ViT-B/32. Ensembling accuracy and language-maxing accuracy in parentheses. Datasets with challenging description behavior - namely Food101, ImageNet, ImageNetV2, and OxfordPets - tend to perform best for high values of $k$. The remaining datasets with favorable description behavior show no uniform pattern but significantly surpass the LLM-assigned baseline. LLM-assigned performance for reference. An evaluation is impossible where $k>|\mathcal{C}|$ (denoted by -). Used parameters: $\text{pool}=\text{DCLIP}$, $m=5$, $n=\text{maximal}$.}
\label{tab:k_eval}
\end{table*}

\begin{table*}[]
\centering\resizebox{\linewidth}{!}{
\begin{tabular}{c|ccccccccc}
\toprule
\textbf{ViT-L/14}&&&&&\textbf{Test Dataset}\\
\midrule
\textbf{$k$}  & CUB200 & DTD & EuroSAT & Flowers102 & Food101 & ImageNet & ImageNetV2 & Oxford Pets & Places365 \\
\midrule
2 & 67.28 (\underline{67.29}) & 58.51 (57.71) & 59.20 (57.81) & 78.83 (77.93) & 90.07 (\underline{90.30}) & 74.16 (\underline{74.38}) & 68.27 (\underline{68.72}) & 91.88 (92.53) & 43.68 (43.58) \\
3 & \textbf{67.81} (67.09) & 58.83 (57.87) & 60.56 (56.56) & 79.69 (79.53) & 90.11 (89.85) & 74.36 (74.28) & 68.15 (68.32) & \textbf{93.05} (\underline{92.61}) & 44.16 (44.03) \\
4 & 67.02 (66.86) & 59.26 (58.56) & 62.26 (58.04) & 80.40 (80.14) & 90.19 (89.66) & 74.32 (74.29) & 68.55 (68.20) & 92.83 (92.20) & 44.48 (\underline{44.08}) \\
5 & 66.83 (66.93) & 60.27 (58.88) & 62.81 (58.61) & 80.87 (80.39) & 90.10 (89.68) & 74.55 (74.22) & 68.40 (68.18) & 92.12 (91.93) & 44.46 (43.99) \\
6 & 66.52 (66.10) &59.79 (58.99) & 63.46 (59.69) & 81.40 (80.52) & 90.19 (89.57) & 74.49 (74.04) & 68.79 (68.36) & 91.85 (91.50) & 44.45 (43.79) \\
7 & 67.40 (66.09) & 60.16 (59.20) & 63.46 (60.86) & 82.01 (80.94) & 90.26 (89.62) & 74.54 (74.06) & 68.30 (68.33) & 91.50 (91.14) & 44.48 (43.80) \\
8 & 66.97 (66.19) & 60.05 (59.20) & 64.73 (64.83) & 82.29 (81.28) & 90.22 (89.77) & 74.49 (73.95) & 68.33 (68.35) & 91.17 (90.76) & 44.44 (43.58) \\
9 & 66.62 (65.57) & 61.12 (58.99) & 65.57 (\underline{66.26}) & 82.66 (81.53) & 90.17 (89.81) & 74.57 (73.97) & 68.62 (68.00) & 90.62 (90.24) & 44.52 (43.67) \\
10 & 66.33 (65.57) & 61.01 (58.94) & \textbf{66.53} (65.51) & \textbf{82.76} (\underline{82.13}) & 90.21 (89.68) & 74.57 (73.97) & 68.40 (68.01) & 90.65 (89.75) & 44.62 (43.59) \\
15 & 66.17 (64.83) & \textbf{61.70} (\underline{60.05}) & - & 82.48 (82.05) & \textbf{90.37} (89.64) & \textbf{74.58} (73.93) & 68.56 (67.73) & 90.62 (89.75) & \textbf{44.65} (43.41) \\
20 & 65.34 (63.69) & 61.22 (58.62) & - & 81.05 (80.53) & 90.28 (89.37) & 74.56 (73.76) & 68.42 (67.64) & 90.71 (89.94) & 44.47 (43.32) \\
25 & 65.43 (63.41) & 61.17 (58.56) & - & 81.10 (80.39) & 90.04 (89.19) & 74.63 (73.75) & \textbf{68.63} (67.69) & 91.36 (90.46) & 44.32 (42.98) \\
30 & 65.14 (63.24) & 61.01 (57.66) & - & 81.07 (80.61) & 89.85 (88.99) & 74.56 (73.53) & 68.61 (67.49) & 90.92 (90.13) & 44.17 (42.81) \\
\midrule
LLM Assignments & 63.39 & 54.73 & 57.97 & 76.83 & 89.94 & 74.47 & 68.74 & 93.51 &   41.66 \\ 
\midrule
Maximum Gain in pp. & \textcolor{green}{4.42} & \textcolor{green}{6.97} & \textcolor{green}{8.56} & \textcolor{green}{5.93} & \textcolor{green}{0.43} & \textcolor{green}{0.09} & \textcolor{red}{-0.02}  & \textcolor{red}{-0.46} & \textcolor{green}{2.99}  \\

\bottomrule

\end{tabular}
}
\caption{Ablation of $k$ for the selected assignments in the classname-containing setup for ViT-L/14. Ensembling accuracy and language-maxing accuracy in parentheses. Datasets with challenging description behavior - namely Food101, ImageNet, ImageNetV2, and OxfordPets - tend to perform best for high values of $k$ while performing on par with the LLM-assigned baseline. The remaining datasets with favorable description behavior show no uniform pattern but significantly surpass the LLM-assigned baseline. An evaluation is impossible where $k>|\mathcal{C}|$ (denoted by -). Used parameters: $\text{pool}=\text{DCLIP}$, $m=5$, $n=\text{maximal}$.}
\label{tab:k_eval_L_bbn}
\end{table*}

\subsection{Evaluating Beyond Ensembling}
\label{subsubsec:max_vs_mean}
Knowing that ensemble effects can boost classification accuracy with non-semantic descriptions, such as random strings, that serve to obtain alternative classname embeddings, an interesting ablation involves eliminating the ensembling from the evaluation process. The ensembling evaluation of image $x$ worked via $\Tilde{c}=\mathop{\argmax}_{c \in \mathcal{C}} s(c,x)$ with $s(c,x) = \frac{1}{|\mathcal{D}(c)|} \sum_{d \in \mathcal{D}(c)} \phi(e(d), e(x))$. The ensembling can be eliminated by using $\Tilde{c}=\mathop{\argmax}_{c \in \mathcal{C}} s(c,x)$ as before but with $s(c,x) = \max_{d \in \mathcal{D}(c)} \phi(e(d), e(x))$ which denotes the maximum image-description similarity of candidate class $c$. This way, no smoothing averaging operation is involved. A maximum operator replaces it. Hence, it can be conjectured that this evaluation procedure is semantics-sensitive, provided that the vision-language embedding space correctly embeds textual description semantics.

\Cref{tab:max_vs_mean} compares the results of both evaluation modes within the DCLIP description pool. It shows that within the DCLIP description pool, the baseline - Ensembling of classwise DCLIP LLM assignments - is \textit{only} exceeded by the selected assignments of the proposed method. The gains are considerable, reaching from $\textcolor{green}{+3.15}$ to up to $\textcolor{green}{+9.81}$. These results are achieved, although the maxing evaluation is ensembling-free. Hence, no use can be made of ensembling effects that boost performances independently of description semantics. On the other hand, when applying the maxing evaluation protocol to the LLM-assigned descriptions, the classification accuracy drops for $6/7$ datasets investigated. Focusing on the highest activating \texttt{``\{cls\},\{description\}"}-embedding hampers the classification accuracy when the assignments \texttt{cls}$ \leftrightarrow $\texttt{description} are obtained by an LLM. In contrast, if these assignments are obtained from the proposed selection method, a considerable performance boost to the ensembling baseline can be observed for $5/7$ datasets. For both datasets, ImageNet and ImageNetV2, no performance boost can be observed when maxing. This also holds for the LLM-assigned descriptions and points to the fact that ImageNet and ImageNetV2 show less sensitivity to description semantics.

\begin{table*}[]
\centering
\resizebox{\linewidth}{!}{
\begin{tabular}{lll|lllllll}
\toprule
 \textbf{Pool} & \textbf{Assignment} & \textbf{Evaluation} & \textbf{ImageNet} & \textbf{ImageNetV2} & \textbf{CUB200} & \textbf{EuroSAT} & \textbf{Places365} & \textbf{DTD} & \textbf{Flowers102}\\ 
\midrule
DCLIP & LLM & Ensembling & \textbf{55.82}  & \textbf{63.12}  & 52.47  &43.29  &40.47  & 43.99 & 64.01\\ 
\midrule
DCLIP & LLM & Maxing & 54.41 \textcolor{red}{(-1.41)}& 61.67 \textcolor{red}{(-1.45)}& 52.40 \textcolor{red}{(-0.07)} & 43.29 (0) & 37.21 \textcolor{red}{(-3.26)} & 43.35 \textcolor{red}{(-0.63)} & 63.62 \textcolor{red}{(-0.39)}\\
DCLIP & Selected & Maxing & 54.42 \textcolor{red}{(-1.40)}& 62.00 \textcolor{red}{(-1.12)}& \textbf{55.62} \textcolor{green}{(3.15)} & \textbf{53.10} \textcolor{green}{(9.81)} & \textbf{42.49} \textcolor{green}{(2.02)} & \textbf{48.08} \textcolor{green}{(4.09)} & \textbf{68.28} \textcolor{green}{(4.27)}\\ 
\bottomrule
\end{tabular}
}
\caption{Comparison of semantic-sensitive maxing evaluation for both assignment types within the DCLIP description pool for ViT-B/32. Evaluation happens in the classname-containing, conventional scenario to compare the classname ensembling effect to the non-ensembling maxing evaluation. Ensembling of LLM-assigned descriptions is referenced as a baseline. Colored differences between maxing evaluation and baseline performance are displayed. For LLM-assigned descriptions, maxing performance falls short of the ensembling baseline performance in $6/7$ cases and exceeds it in no case. Contrary to this, the maxing evaluation for the selected assignments exceeds ensembling baseline performance in $5/7$ cases with considerable gains. Thus, the enhanced semantic distinctiveness of the selected descriptions enables to surpass the classname ensembling effect. } 
\label{tab:max_vs_mean}
\end{table*}

\subsection{Construction of the description pool \texorpdfstring{$\mathcal{P}$}{P} from existing method.} 
\label{app:construction_pool}
To construct the description pool $\mathcal{P}$, one source is to utilize published descriptions from other works, such as DCLIP~\cite{dclip}. However, since DCLIP does not provide descriptions for Flowers102, we generated descriptions for this dataset using GPT-3.5 with prompts from their code base. In these scenarios, the LLM "assigns" descriptions to a class by generating them specifically for that class. Therefore, both LLM assignments and our assignments, as described in ~\cref{alg:pseudocode}, are included in the DCLIP description pool. In rare instances where the LLM did not return a description following the DCLIP approach, a neutral description, such as "a kind of food" for the food dataset, was used. This step was necessary to avoid distorting the results in this setup otherwise.

\noindent\paragraph{Contrastive prompting} As the vanilla CLIP model can locate the image embedding in an approximate correct neighborhood already with only \texttt{cls} provided, we can use this information to find out the classes that are usually misclassified to each other. Instead of defining ambiguous classes for each image, we obtain the contrastive description pool with a statistical sample of ambiguous classes for each class $c \in  \mathcal{C}$. All the pairs $\forall c \in \mathcal{C}: \forall a \in \mathcal{A}(c): (c,a)$, where we know which classes are usually miss-classified by CLIP solely based on \texttt{cls}. This information is then used to prompt an LLM to generate distinguishing attributes between $c$ and $a$. For each class in its both possible roles $c$ and $a$, all the generated descriptions obtained in this way get collected. This yields an LLM-assignment $\forall c \in \mathcal{C}: \mathcal{D}(c)_{LLM}$. The LLM assignments can be removed to obtain a global description pool $\mathcal{P}$ without any assignments $\mathcal{D}(c)$. The global pool then allows the application of~\cref {alg:pseudocode}, which yields class assignments $\mathcal{D}(c)$ per image $x_i$ based on feedback from the VLM embedding space. \\

\subsection{Comparison to prior works in the classname-free setting}
\label{subsec:app:classname-free_compare}

As briefly discussed in the main text, we are the first to search and evaluate distinctive classname-free textual descriptions in a VLM ensembling scenario. However, some studies have employed trainable bottlenecks or distorted the underlying embedding space in classname-free scenarios. The differences are discussed below:
\begin{itemize}
    \item Works like \textit{LaBo}~\cite{labo} use large Bottleneck sizes (from 500 for CUB up to 50,000 for ImageNet. They are mostly larger than the original CLIP embedding dimensionality, cf. Table 16 in the appendix of \citet{labo}. Therefore, the CLIP image embedding gets transformed to an overcomplete basis and overlaps 100\% between classes. This requires that several thousands of continuous linear classifier weights be interpreted for every class, which is quite cumbersome.
    \item In the case of \textit{Concise and descriptive Descriptions}~\cite{concise_and_descriptive}, the Bottleneck size is much smaller than in Labo, \eg 64. However, the Bottleneck is still shared totally among all the classes, and one has to interpret a vector of 64 continuous weights per class.
    \item In the case of \textit{Pre-trained vision-language models learn discoverable visual concepts}~\cite{zang2024pretrained}, the bottleneck sizes are quite large, though the associated weights are more interpretable as they are only elements of ${0,1}$. However, the text and image embeddings both get projected through a linear layer first, thus leaving the original, generally valid CLIP space. This raises questions about the general validity of the used image-language similarities. 
\end{itemize}

\noindent\citet{vltaboo} evaluated VLM classification by relying only on language descriptions but did not analyze the interplay between \textit{separate} descriptions and a simultaneously introduced classname prompt $d^{c+}$, which is crucial to understand the resulting accuracy in VLM ensembling scenarios. In addition to that, every image description was appended to the same text prompt. The resulting long prompts degraded performance.

\subsection{Chosen selection sizes \texorpdfstring{$n$}{n}}
\label{app:hyperparameter_n}
Per dataset, the sample sizes $n$ that are displayed in~\Cref{tab:hyperparameter_n} were utilized for the experiments of~\Cref{tab:class_name_less} and ~\Cref{tab:main_convention}.

\begin{table}[h]
\centering
\begin{tabular}{l|c}
\toprule
\textbf{\textbf{Dataset}} & \textbf{n} \\
\midrule
CUB 200  & 29 \\

DTD  & 40 \\

Eurosat & 1000 \\

Flowers 102 & 20 \\

ImageNet & 732 \\

ImageNetV2 & 732  \\

\bottomrule
\end{tabular}

\caption{Number of Image Samples per class $n$ used to construct the selection matrix $S$. This number corresponds to the smallest cardinality of all classes in the respective train sets, \ie $n=\mathop{\min}_{c \in \mathcal{C}} |c_{\text{train}}|$. For Eurosat it was arbitrarily set to 1000 because no train split was provided. 
}
\label{tab:hyperparameter_n}
\end{table}

\subsection{Varying the LLM.}
\label{app:llm_backbone}

\Cref{tab:llm_backbone} demonstrates that our algorithm functions effectively with contrastively obtained description pools from both LLMs LlaMa and ChatGPT. This illustrates the versatility of our method.
\begin{table}[]
\centering
\resizebox{\linewidth}{!}{
\begin{tabular}{l|cccccccc}
\toprule
\textbf{LLMs}              & \textbf{ImageNet} & \textbf{ImageNetV2} & \textbf{CUB200} & \textbf{EuroSAT} & \textbf{Places365} & \textbf{DTD} & \textbf{Flowers102}\\
\midrule
Contrastive Llama3-70B   & 61.68 & 54.64 &53.49 & 36.33 & 41.33 & 79.87 & 64.45 \\

w/ \textit{Our Selection}                & \textbf{63.30} & \textbf{55.24}& \textbf{56.27} & \textbf{58.57} & \textbf{43.65}  & \textbf{48.09} & \textbf{68.61} \\

\midrule

Contrastive GPT3.5          & 61.65 & 54.69 & 53.83  & 35.38  & 40.79    & 44.10 & 66.29 \\

w/ \textit{Our Selection}                & \textbf{63.15} & \textbf{55.09} & \textbf{56.21}  & \textbf{44.43}  & \textbf{43.64}  & \textbf{80.78} & \textbf{68.48}\\

\bottomrule
\end{tabular}
}
\caption{Performance of classname-included descriptions: LLM assignments vs Our selection assignments.
}
\label{tab:llm_backbone}
\end{table}

\subsection{Varying the VLM.}
\label{app:vlm_backbone}
\Cref{tab:vlm_backbone} demonstrates that our algorithm functions effectively within the ALIGN VLM embedding space~\cite{jia2021ALIGN}. This further illustrates the versatility of our method.

\begin{table}[]
\centering
\resizebox{\linewidth}{!}{
\begin{tabular}{l|cccccccc}
\toprule
\textbf{ALIGN}              & \textbf{ImageNet} & \textbf{ImageNetV2} & \textbf{CUB200} & \textbf{EuroSAT} & \textbf{Places365} & \textbf{DTD} & \textbf{Flowers102}\\
\midrule
ALIGN & 59.83 &	54.59 &	37.54 &	29.62 &	39.89 &	55.00 &	54.26 \\

+LLM assigned & 59.87 &	54.84 &	37.25 &	29.37 &	40.24 &	57.29 &	55.28 \\

+randomly assigned  & 59.83 & 54.57 &	37.14 &	29.14 &	39.89 &	57.18 &	55.46 \\

w/ \textit{Our Selection}   & \textbf{60.26} & \textbf{55.29} & \textbf{40.32} & \textbf{31.74} & \textbf{42.47} & \textbf{59.79} &\textbf{55.91}\\

\bottomrule
\end{tabular}
}
\caption{Performance in the classname-free evaluation setup using the ALIGN VLM.}
\label{tab:vlm_backbone}
\end{table}

\subsection{Examples of selected descriptions}
\label{subsec:app:example_descriptions}
In this section, we show the obtained descriptions for different datasets in the \textit{json} style below. The ambiguous classes and corresponding selected descriptions of two images for several datasets are shown.
\newpage

\begin{lstlisting}[caption={Examples of selected descriptions by our algorithm}]
{
 "image_79": {
            "highway or road": [
                "surrounded by parking lots or roads",
                "parking spaces or driveways",
                "man-made structures like lamps or signs",
                "road markings",
                "vehicles"
            ],
            "brushland or shrubland": [
                "dense vegetation",
                "tree trunks and branches",
                "random patterns",
                "natural curves and lines",
                "randomly distributed shadows"
            ],
            "permanent crop land": [
                "change in crop type or growth stage through the seasons",
                "green color during growing season",
                "texture of crops such as rows of plants or scattered crops",
                "farmland patterns",
                "irregular shapes"
            ]
        },
"image_13182": {
            "residential buildings or homes or apartments": [
                "uninterrupted spaces",
                "straight rows",
                "man-made structures"
            ],
            "industrial buildings or commercial buildings": [
                "may be clustered together in groups"
            ],
            "permanent crop land": [
                "large, rectangular fields",
                "straight rows of crops",
                "irrigation systems",
                "green vegetation",
                "typically have a rectangular or square shape"
            ]
        },
"image_1103": {
            "perforated": [
                "Intricate and delicate patterns",
                "repeated floral or geometric design",
                "Raised ridges on the surface",
                "Uniform pattern of parallel ridges",
                "consistent patterns"
            ],
            "crosshatched": [
                "often darker color",
                "stretchy and flexible",
                "often surrounded by loose debris",
                "often darker coloration",
                "irregular swirls or streaks"
            ],
            "meshed": [
                "fine mesh-like structure",
                "straight lines",
                "sharp lines",
                "continuous curved lines",
                "distinct layers of different colors"
            ]
        },
"image_1440": {
            "smeared": [
                "intermingling of colors",
                "consisting of straight lines",
                "uneven distribution of color",
                "seemingly messy look",
                "blurred edges"
            ],
            "stained": [
                "fine, thin strands",
                "stretchy and flexible",
                "Curved teardrop shapes",
                "long and fine strands",
                "Folds of fabric that create a wavy pattern"
            ],
            "blotchy": [
                "Intricate and delicate patterns",
                "fine mesh-like structure",
                "repeated floral or geometric design",
                "Raised ridges on the surface",
                "Uniform pattern of parallel ridges"
            ]
        },
"image_4747": {
            "Chestnut sided Warbler": [
                "yellow face",
                "greenish upperparts",
                "white underparts",
                "split white wing bars",
                "white spectacles"
            ],
            "Nelson Sharp tailed Sparrow": [
                "Bay breast and flanks",
                "black streaking on sides",
                "black face",
                "bold black streaking on sides",
                "black streaking on its back and sides"
            ],
            "Bay breasted Warbler": [
                "chestnut-colored flanks",
                "distinctive chestnut-colored patch on its flanks",
                "chestnut-colored crown",
                "Black cap",
                "olive-green upperparts"
            ]
        },
"image_3485": {
            "Eastern Towhee": [
                "Bay breast and flanks",
                "black streaking on sides",
                "greenish upperparts",
                "black face",
                "chestnut-colored flanks"
            ],
            "Harris Sparrow": [
                "distinctive chestnut-colored patch on its flanks",
                "white underparts",
                "pale underparts",
                "Pale underparts",
                "grayish-brown overall coloration"
            ],
            "Northern Waterthrush": [
                "yellow face",
                "olive-green upperparts",
                "grayish-olive upperparts",
                "white eye ring",
                "small size"
            ]
        },
"image_4716": {
            "clematis": [
                "lance-shaped leaves",
                "daisy-like flowers",
                "single row of petals",
                "stem is hairy",
                " colorful flowers, but not bright red"
            ],
            "columbine": [
                " cluster of stems",
                "daisy-like flowers with white petals and yellow centers",
                "delicate petals",
                "upright, clump-forming habit",
                "pale blue, white, or pink flowers"
            ],
            "balloon flower": [
                "leaves are simple and alternate",
                "cup-shaped flowers",
                "tall stem",
                " solitary stem",
                "umbrella-like leaf structure"
            ]
        },
"image_4981": {
            "desert-rose": [
                " cluster of stems",
                " colorful flowers, but not bright red",
                "thick, waxy leaves",
                "typically rosette-forming growth habit",
                "fragrant, delicate, pale pink to deep pink flowers"
            ],
            "frangipani": [
                "lance-shaped leaves",
                "daisy-like flowers with white petals and yellow centers",
                "daisy-like flowers",
                "leaves are simple and alternate",
                "cup-shaped flowers"
            ],
            "mexican petunia": [
                "single row of petals",
                "stem is hairy",
                "delicate petals",
                "tall stem",
                " solitary stem"
            ]
        },
"image_18780": {
            "ski_slope": [
                "icy slopes",
                "ski lifts",
                "ski slopes",
                "ski equipment",
                "ski tracks"
            ],
            "bridge": [
                "ornamental bridges",
                "arches or spans for carrying a road or railway",
                "river or stream",
                "natural rock formations around the water",
                "surrounded by walls or embankments"
            ],
            "ski_resort": [
                "gambrel or gable roof",
                "snow cover",
                "covered in snow",
                "snow-capped roof",
                "snow-covered roof"
            ]
        },
"image_100": {
            "banquet_hall": [
                "seating arrangement for events",
                "chairs arranged for dining",
                "decorative table settings",
                "place settings on the table",
                "formal table settings"
            ],
            "dining_hall": [
                "outdoor seating area",
                "outdoor seating",
                "patio seating",
                "grilling area",
                "food served in containers"
            ],
            "ballroom": [
                "dancing",
                "performer on a central stage",
                "vertical sliding movement",
                "skaters wearing figure skates",
                "steps"
            ]
        },
"image_20": {
            "Abyssinian": [
                "almond-shaped eyes",
                "dark, expressive eyes",
                "large eyes",
                "large, expressive eyes",
                "soulful eyes"
            ],
            "Bengal": [
                "wrinkled face",
                "curled tail",
                "thick, fluffy coat",
                "high energy level",
                "bushy eyebrows and beard"
            ],
            "Egyptian Mau": [
                "a long, thick coat that is usually white with darker markings",
                "a thick, double coat of fur that is black and silver or black and cream in color",
                "blue-grey fur",
                "white markings on the chest, feet, and face",
                "black, fawn, or silver coat"
            ]
        },
"image_2186": {
            "newfoundland": [
                "black, grey, or brown fur",
                "dark brown or black fur",
                "brindle, black, or blue coat",
                "brindle, fawn, or black coat",
                "black, grey, or brindle"
            ],
            "great pyrenees": [
                "a long, thick coat that is usually white with darker markings",
                "a large, white, fluffy dog",
                "a dense, wavy coat that is wheaten in color",
                "a white, fluffy coat",
                "white paws"
            ],
            "english cocker spaniel": [
                "droopy ears",
                "wet nose",
                "webbed feet",
                "small ears",
                "long, droopy ears"
            ]
        },
}
\end{lstlisting}

\subsection{Example of a description pool P}
\label{subsec:app:example_pool}
In this section, we show parts of the LLM assignments of the places dataset. Dissolving the LLM assignments and collecting the classname-less descriptions in a global pool yields the pool $P$. It is such a pool $P$ that our \Cref{alg:pseudocode} selects from. \\

\begin{lstlisting}
{
    "index_to_descriptions": {
        "0": [
            "a repeating pattern of light and dark bands",
            "the bands are of different widths",
            "the bands may be of different colors",
            "the bands may be curved or straight",
            "the bands may be parallel or intersecting"
        ],
        "1": [
            "an uneven or mottled surface",
            "a variety of colors or shades",
            "a raised or bumpy texture",
            "a matte finish"
        ],
        "2": [
            "three or more strands of material woven together",
            "a tight, interlocking pattern"
        ],
        "3": [
            "small, round, and raised bumps",
            "a smooth or glossy surface",
            "a three-dimensional appearance",
            "a light-reflecting quality"
        ],
        "4": [
            "an uneven surface",
            "raised or indented areas",
            "a rough or bumpy feel"
        ],
        "5": [
            "a repeating pattern of squares or rectangles",
            "alternating light and dark colors",
            "sharp, defined lines between the squares or rectangles"
        ],
        "6": [
            "a web-like pattern",
            "made of thin, silky strands",
            "often found in dark, damp places",
            "can be sticky to the touch",
            "can be difficult to remove once entangled"
        ],
        "7": [
            "a surface with cracks",
            "the cracks may be straight or curved",
            "the cracks may be of different sizes",
            "the cracks may be close together or far apart",
            "the cracks may be deep or shallow",
            "the cracks may be filled with dirt or debris"
        ],
        "8": [
            "a series of parallel lines that intersect to form a grid",
            "the lines may be of different thicknesses",
            "the lines may be of different colors",
            "the texture may be regular or irregular",
            "the texture may be applied to a surface or object"
        ],
        "9": [
            "a repeating pattern of shapes",
            "sharp edges",
            "a glossy or shiny surface",
            "a transparent or translucent appearance",
            "a three-dimensional structure"
        ],
        "10": [
            "a series of small, round dots",
            "evenly spaced",
            "can be of any color",
            "may be on a background of any color",
            "may be in a regular or irregular pattern"
        ],
        .
        .
        .
    },
    "index_to_classname": {
        "0": "banded",
        "1": "blotchy",
        "2": "braided",
        "3": "bubbly",
        "4": "bumpy",
        "5": "chequered",
        "6": "cobwebbed",
        "7": "cracked",
        "8": "crosshatched",
        "9": "crystalline",
        "10": "dotted",
        "11": "fibrous",
        .
        .
        .
    }
}
\end{lstlisting}

\subsection{Distribution of Distinctiveness Scores}
\label{app:distinctiveness_scores}

\begin{figure*}
\centering
    \begin{subfigure}{0.4\textwidth}
        \includegraphics[width=\linewidth]{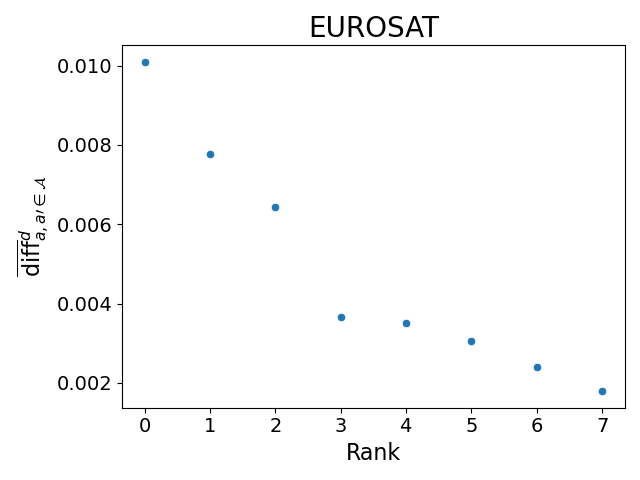}
    \end{subfigure}
    \begin{subfigure}{0.4\textwidth}
        \includegraphics[width=\linewidth]{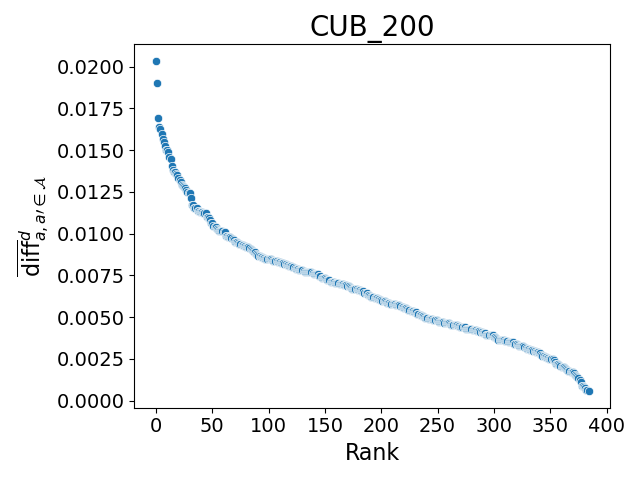}
    \end{subfigure}
    \begin{subfigure}{0.4\textwidth}
        \includegraphics[width=\linewidth]{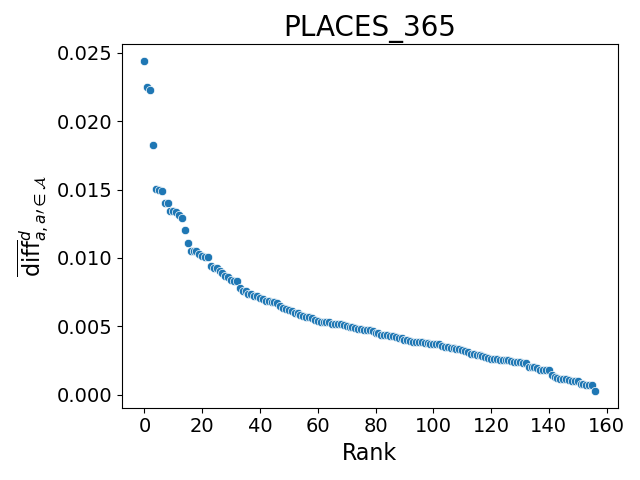}
    \end{subfigure}
        \begin{subfigure}{0.4\textwidth}
        \includegraphics[width=\linewidth]{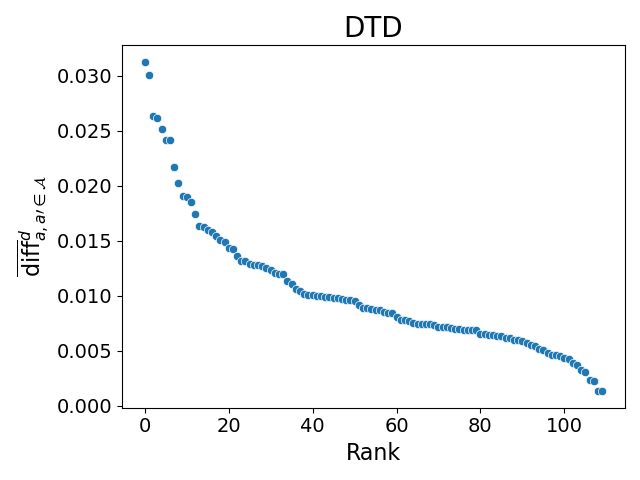}
    \end{subfigure}
    \begin{subfigure}{0.4\textwidth}
        \includegraphics[width=\linewidth]{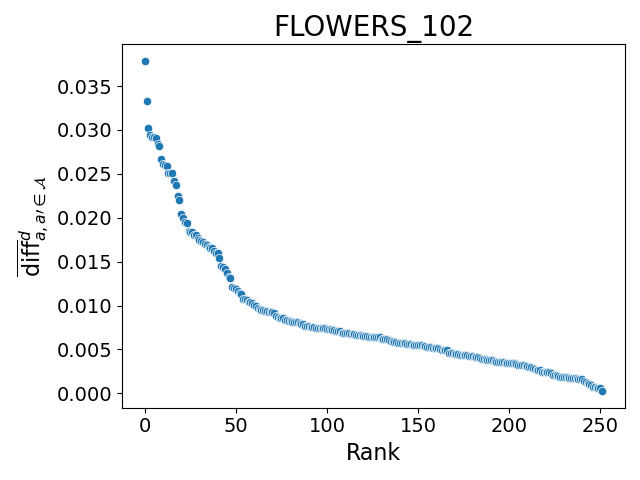}
    \end{subfigure}
    \begin{subfigure}{0.4\textwidth}
        \includegraphics[width=\linewidth]{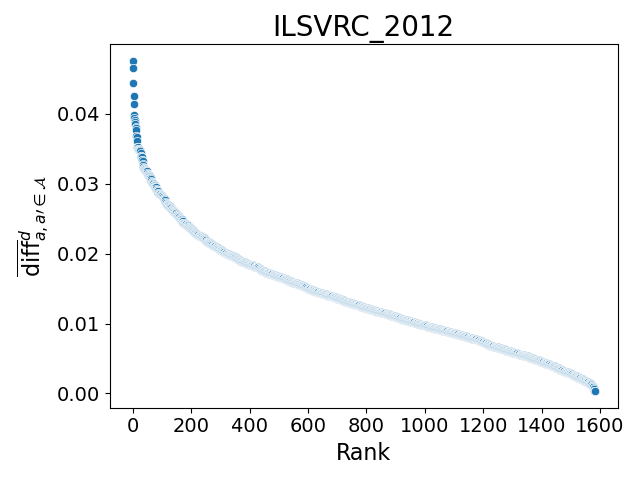}
    \end{subfigure}
    \begin{subfigure}{0.4\textwidth}
        \includegraphics[width=\linewidth]{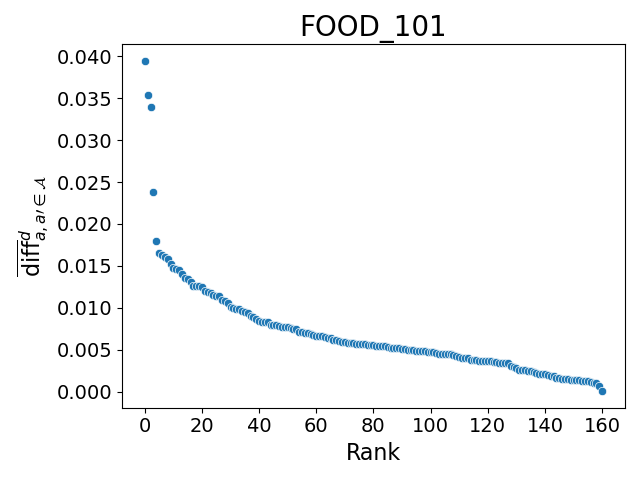}
    \end{subfigure}  
    \begin{subfigure}{0.4\textwidth}
        \includegraphics[width=\linewidth]{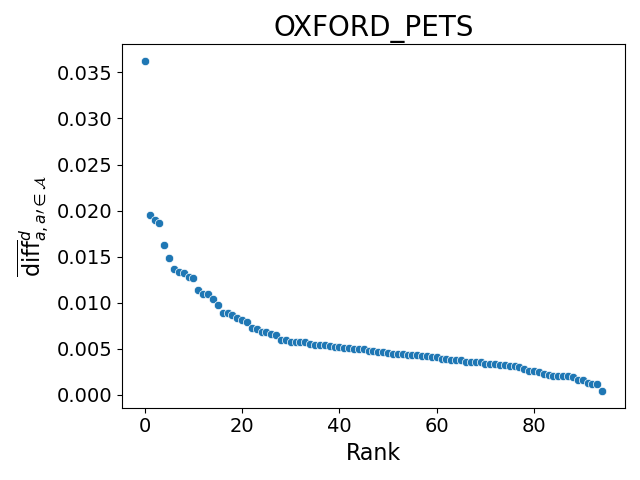}
    \end{subfigure}  
    \caption{Distinctiveness scores of randomly chosen images obtained by the training-free approach presented in~\Cref{sec:vlm_feedback}. Distinctiveness scores $\overline{\mathrm{diff}}_{a,a\prime\in \mathcal{A}}^d =\frac{1}{k-1} \sum_{a\prime\in \mathcal{A}} \mathrm{diff}_{a,a'}^d=\bar{s}_{a,d}-\bar{s}_{a',d}$ where $ \mathrm{diff}_{a,a'}^d=\bar{s}_{a,d}-\bar{s}_{a',d} \geq 0$. Used Parameters: $k=3$, $m=5$, $n=\text{maximal}$, pool = DCLIP. See~\Cref{app:distinctiveness_scores} for a concise discussion.} 

\label{fig:distinctiveness_scores_appendix}
\end{figure*}

\Cref{fig:distinctiveness_scores_appendix} shows the ranked distribution of distinctiveness scores obtained by the training-free method in $\emph{S}^{+}_i$ of~\Cref{alg:pseudocode}. For one randomly chosen image per dataset and one randomly chosen ambiguous class, all positive values of $\overline{\mathrm{diff}}_{a,a\prime\in \mathcal{A}}^d$ are displayed, sorted by their rank. These values correspond to the distinctiveness of descriptions in $S^+$. Notably, the ranked distribution of the distinctiveness scores appears to follow a Pareto distribution: A few top-ranked descriptions score substantially higher than the rest, while most descriptions score significantly lower and are closer to each other, which is characteristic of a Pareto distribution. These highest-scoring descriptions offer the highest distinctiveness according to the available samples in $S$ \wrt $a$. Since only relatively few descriptions yield high distinctiveness scores, this concise set of highly distinctive descriptions can be captured by a low value of $m$. Thus, selecting the top-$m$ scoring descriptions via $m=5$, already captures large parts of the steeply declining highly distinctive descriptions. This explains why already $5$ selected descriptions in the DCLIP description pool bring substantial performance gains, as seen in~\Cref{tab:main_convention} and~\Cref{tab:main_vit_l14}.

\end{document}